\documentclass[10pt,twocolumn,letterpaper]{article}

\PassOptionsToPackage{table,dvipsnames}{xcolor}

\usepackage[iccvfinal]{iccv}

\usepackage[pagebackref,breaklinks,colorlinks,allcolors=iccvblue]{hyperref}

\usepackage{url}
\usepackage{graphicx}
\usepackage{pifont}
\usepackage{amssymb}
\usepackage{makecell}
\usepackage{multirow}
\usepackage{caption}
\usepackage{booktabs}
\usepackage{varwidth}
\usepackage{xspace}
\usepackage{wrapfig,lipsum}

\definecolor{iccvblue}{rgb}{0.21,0.49,0.74}

\def\paperID{12262} 
\def\confName{ICCV}
\def\confYear{2025}
\newcommand{\method}{\textsc{VIM}\xspace}
\newcommand{\methodbold}{\textbf{\textsc{VIM}\xspace}}
\newcommand{\dataset}{\textsc{$\text{I}$}nter-\textsc{$\text{MT}^2$}\xspace}
\newcommand{\datasetbold}{\textbf{\textsc{$\text{I}$}nter-\textsc{$\text{MT}^2$}}\xspace}

\newcommand{\vio}[1]{{\color{violet}\small {\sf{#1}}}}

\definecolor{lightgray}{gray}{0.90}
\newcommand\greybox[1]{%
  \vskip\baselineskip%
  \par\noindent\colorbox{lightgray}{%
    \begin{minipage}{0.98\linewidth} {\color{gray} \small \sf{#1}}\end{minipage}%
  }%
  \vskip\baselineskip%
  
}
\newcommand\blfootnote[1]{%
  \begingroup
  \renewcommand\thefootnote{}\footnote{#1}%
  \addtocounter{footnote}{-1}%
  \endgroup
}

\title{A Unified Framework for Motion Reasoning and Generation \\ in Human Interaction}

\author{Jeongeun Park$^{1}$\thanks{}\qquad
Sungjoon Choi$^{1\dagger}$\qquad 
Sangdoo Yun$^{2\dagger}$ \\
$^{1}$Korea University \quad $^{2}$ NAVER AI Lab\\
\url{https://vim-motion-language.github.io/} \\ 
}

\begin{document}
\twocolumn[{%
\renewcommand\twocolumn[1][]{#1}%
\maketitle
 \vspace*{-1cm}
\begin{center}
    \centering
    \captionsetup{type=figure}
    \includegraphics[width=0.9\linewidth]{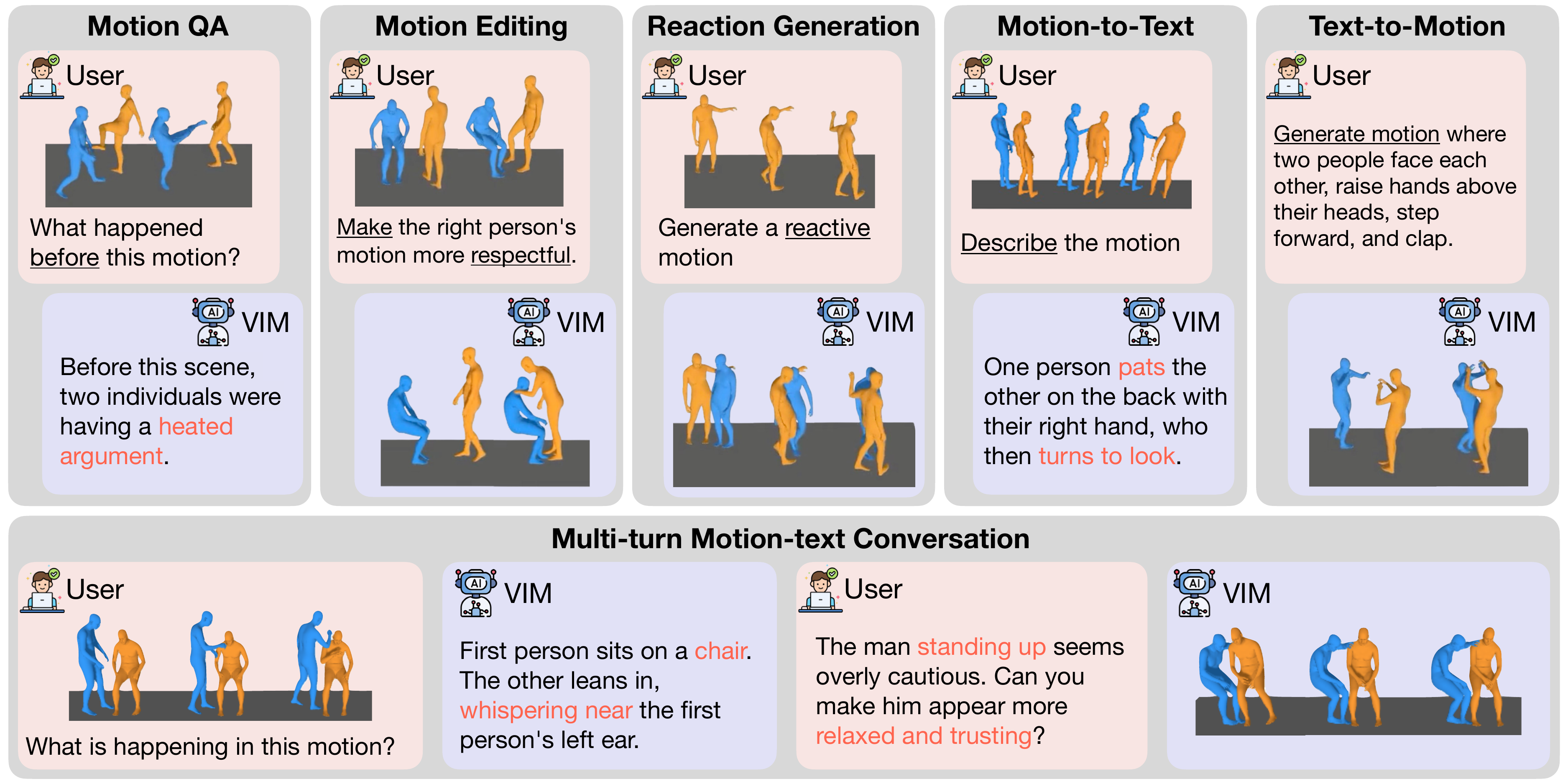}
   \vspace{-0.4em}    
    \captionof{figure}{We introduce \methodbold, the Versatile Interactive Motion-language model, a unified architecture that combines language and motion for two-person interactive scenarios. The figure highlights its capabilities across various tasks including motion-to-text, text-to-motion, reaction generation, motion editing, and multi-turn motion reasoning, all within a single framework. } \label{fig:teaser} 
  \vspace{-0.4em}
\end{center}%
}]

\blfootnote{\hspace{-0.5cm} $^*$ Work done during internship at NAVER AI Lab.  $^\dagger$ Corresponding authors.}

\begin{abstract}
Recent advancements in large language models (LLMs) have greatly enhanced their ability to generate natural and contextually relevant text, enabling more human-like AI interactions. However, generating and understanding interactive human-like motion, where multiple individuals engage in coordinated movements, remains challenging due to the complexity of modeling these coordinated interactions. 
Furthermore, a unified and versatile model is required to handle diverse interactive scenarios, such as chat systems that dynamically adapt to user instructions and assigned roles.
To tackle these problems, we introduce \methodbold, the Versatile Interactive Motion-language model, which integrates both language and motion modalities to effectively understand, generate, and control interactive motions in multi-turn conversational contexts.
Unlike previous studies primarily focusing on uni-directional tasks (e.g., text-to-motion or motion-to-text), \methodbold~employs a unified architecture capable of simultaneously understanding and generating both motion and text modalities. 
Given the lack of an appropriate dataset to address this challenge, we introduce \datasetbold, a large-scale instruction-tuning dataset containing 82.7K multi-turn interactive motion instructions, spanning 153K interactive motion samples. \datasetbold covers diverse instructional scenarios including editing, question answering, and story generation, with interactive motions leveraging off-the-shelf large language models and motion diffusion models.
We extensively evaluate the versatility of \methodbold~across multiple interactive motion-related tasks: motion-to-text, text-to-motion, reaction generation, motion editing, and reasoning about motion sequences. Remarkably, \methodbold~is the first model capable of effectively addressing all these tasks with a single unified framework, achieving competitive performance compared to task-specific methods. 
\end{abstract}    
\section{Introduction}
\label{sec:intro}


Modeling interactive human motions stands at the forefront of advancements in robotics and virtual reality. By capturing the subtle nuances of human communications, including gestures, expressions, and interactive behaviors, machines can offer seamless and natural interfaces. 
This holistic understanding enables technology to adjust its responses and behaviors based on the user's physical motions and situational context, leading to more personalized and engaging interactions. 


Recent advancements in large language models (LLMs) \cite{dubey2024llama, team2024gemma, yang2024qwen2} have demonstrated significant potential in generating human-like text and understanding complex linguistic interactions.
They have even extended their capability to multi-modal contexts, successfully integrating various input sources such as images, speech, and videos \cite{ge2024seed,liu2024visual,chen2024comm, tang2024salmonn, shu2023llasm}.
Building upon these developments, there is a growing interest in incorporating human (or robot) motion as a new modality~\cite{jiang2024motionchain,chen24motionllm}, leading to the emergence of the ``motion-language models" (MLM). However, existing approaches \cite{zhang2023generating,guo2024momask,guo2022tm2t,zhang2024motiongpt,Cai_2024_CVPR} often focus on unidirectional tasks that handle one-way translation between text and motion, e.g., text-to-motion or motion-to-text, and consider only single-person motions without interactions. This limitation hinders the agents' ability to handle scenarios involving interactive motions in multi-turn conversations.

Beyond modeling single-person motions, interactive motions between two individuals allow the model to learn about social behavior. Modeling such interactions requires versatility to effectively control interactions, allowing users to provide instructions, assign roles, or modify behaviors. 
In this paper, we aim to build a unified yet versatile motion-language model designed to generate, control, and comprehend sophisticated interactive motions. 

One of the primary challenges in constructing those models is the lack of multi-turn interactive motion data. Datasets containing motions of two individuals interacting with each other, along with multi-turn conversational instructions, are scarce and challenging to collect. This makes it difficult for models to learn the nuances of interactive motions and multi-turn dynamics.
To address this, we present a new interactive motion dataset, \datasetbold, which contains 82K samples, including various instructional scenarios about the interactive motions in a multi-turn conversational format. We utilize large language models to produce diverse instructions with motion captions and diffusion-based text-to-motion models to generate corresponding interaction motions.

Building upon our \datasetbold, we present \methodbold, a \textbf{V}ersatile \textbf{I}nteractive \textbf{M}otion-language model designed for multi-turn conversations involving interactive motions. 
We pursue the versatility of \method through a \textbf{unified architecture} that can simultaneously input and output both motion and text modalities. Based on the pre-trained LLMs, our training process can be divided into three stages: (1) training of the interactive motion tokenizer, (2) pre-training for motion and text representation alignment, and (3) instruction tuning with \dataset~to handle more complex and multi-turn instructions.
This enables \method~to effectively comprehend, generate, and control interactive motions, as illustrated in Figure \ref{fig:teaser}. To evaluate \method's capabilities, we introduce new protocols that assess its performance on various motion-related tasks, including motion editing and reasoning based on contextual cues, demonstrating its versatility in complex scenarios\footnote{We provide more video demos in the supplementary material.}.
We will publicly release our dataset, codes, and models to facilitate future research.

In summary, the main contributions of this paper are threefold: 
(1) We propose \method that can simultaneously process and generate both two-people motion and text modalities, along with a three-stage training pipeline consisting of motion tokenizer training, pre-training for modality alignment, and instruction tuning.
(2) We present \dataset, a multi-turn interactive motion-text dataset, to address the lack of multi-turn interactive motion data.
(3) We introduce a new evaluation protocol to evaluate the performance of motion-language models on complex motion interaction scenarios. 
\section{Related Work}
\label{sec:formatting}

\paragraph{Human Motion Modeling \& Control}

Advancements in human motion modeling have driven significant progress in motion generation and control. Diffusion-based methods~\cite{tevet2023human, wang2023fg, zhang2024motiondiffuse} have been applied to synthesize human motions from text descriptions. Meanwhile, transformer models using vector quantization~\cite{guo2022tm2t, zhang2023generating} have been explored for capturing diverse motion patterns, and MoMASK~\cite{guo2024momask} uses residual tokenizers to enhance fine-grained motion details.
For motion editing, some approaches focus on style transfer \cite{aberman2020unpaired, guo2024generative} or specific body part modifications \cite{zhang2024motiondiffuse, kim2023flame}. MEOs \cite{goel24meo} use captions and large language models to identify frames and body parts to edit, while MotionFix \cite{athanasiou2024motionfix} conditions diffusion models on both source motion and edit text for seamless motion edits.
However, these models usually target unidirectional tasks (\textit{e.g.}, text-to-motion, or motion editing) and cannot handle input and output of both motion and text simultaneously in a unified architecture. Unlike existing methods, our approach processes both motion and text concurrently in a unified architecture.

\paragraph{Motion-Language Model}
Recent developments in motion-language models have aimed to achieve versatility across various motion-related tasks.
MotionGPT \cite{jiang2023motiongpt} demonstrates versatility in motion comprehension and generation based on a unified framework. MotionChain \cite{jiang2024motionchain} introduces a multi-turn conversational system for interpreting and generating motions within dialogue contexts, including image inputs.
Recent work \cite{Zhou_2024_CVPR, chen24motionllm, zhang2024large, luo2024m} has explored unified approaches to multi-modal motion generation, including speech, video, and image.
However, these methods focus on the single-person motions, thus, modeling \textit{interactive motions} in versatile large models remains under-explored. Wu et al. \cite{wu2024motionllm}, address the interactive motions, but they still lack multi-turn interactions and complex reasoning abilities. 
Our work addresses such issues with a model trained on our  \dataset~dataset, enabling the understanding and generation of interactive motions in multi-turn conversations with advanced reasoning capabilities. 

\paragraph{Human-Human Interactive Motion Modeling}
Modeling human-human interactions has garnered increasing attention in recent research. Several multi-person interaction datasets~\cite{ng2020you2me,fieraru2020three,yin2023hi4d} have been developed, and recent efforts like Inter-X \cite{xu2024inter} and InterHuman \cite{liang2024intergen} have collected interactive motions paired with textual descriptions for text-based motion control.
In text-to-motion tasks~\cite{ponce2024in2in, Cai_2024_CVPR}, InterGEN~\cite{xu2024inter} leverages diffusion with spatial constraint loss, while PriorMDM~\cite{shafir2024human} adapts pre-trained diffusion models with slim communication blocks. For reaction generation, ReMoS~\cite{ghosh2023remos} uses spatio-temporal cross-attention to synthesize reactive motions, and ReGenNet~\cite{Xu_2024_CVPR} predicts reactions with a transformer and relative distance-based interaction loss.
While existing models have advanced interactive motion modeling, they lack versatility and focus on specific tasks, failing to capture complex multi-turn dynamics. To address this, we introduce \dataset, enabling agents to generate sophisticated motions, respond to instructions, adapt roles, and adjust behaviors based on context.  


\section{$\text{Inter}\text{-MT}^2$: Interactive multi-turn motion-text dataset }
\begin{figure}
    \centering
    \begin{subfigure}{0.5\textwidth}
        \includegraphics[width=\textwidth]{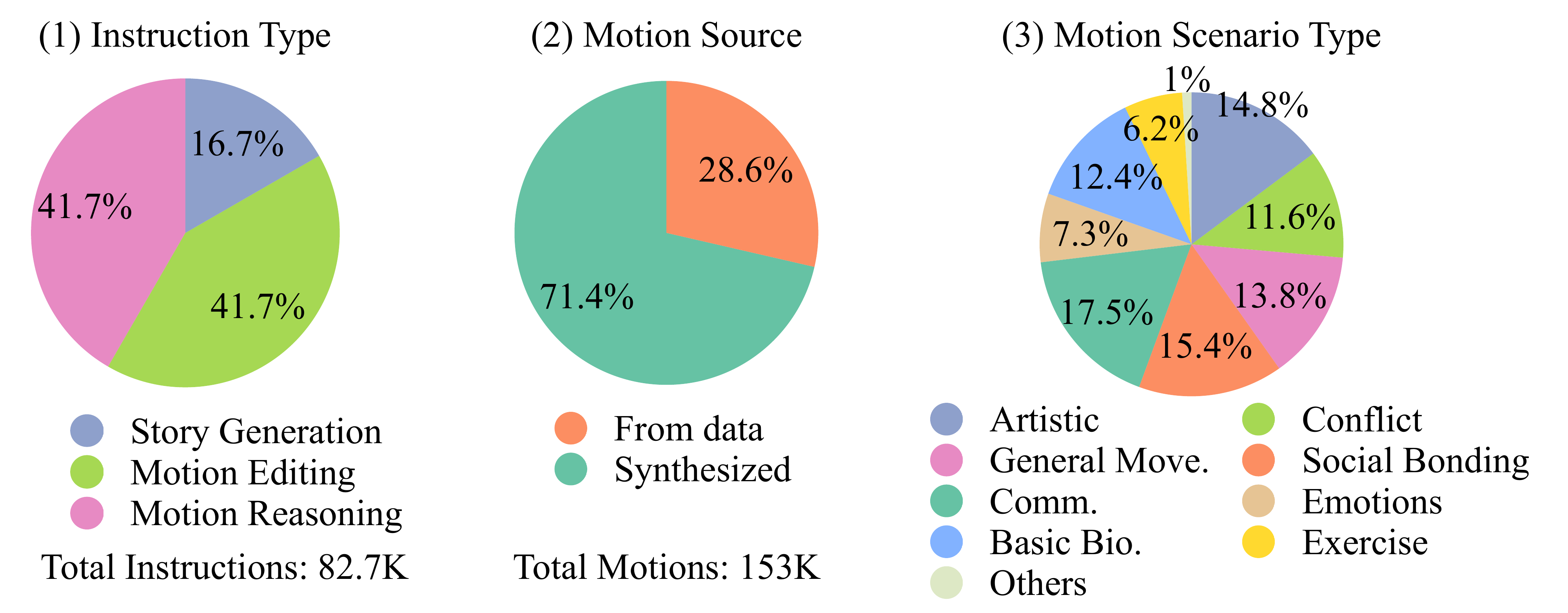}
    \caption{The distribution of instruction types, motion sources, and motion scenario types, highlighting the dataset's diversity. The type of motion scenario is classified using a large language model with motion captions.}
    \label{fig:short-a}
    \end{subfigure}
    \begin{subfigure}{0.5\textwidth}
        \includegraphics[width=\textwidth]{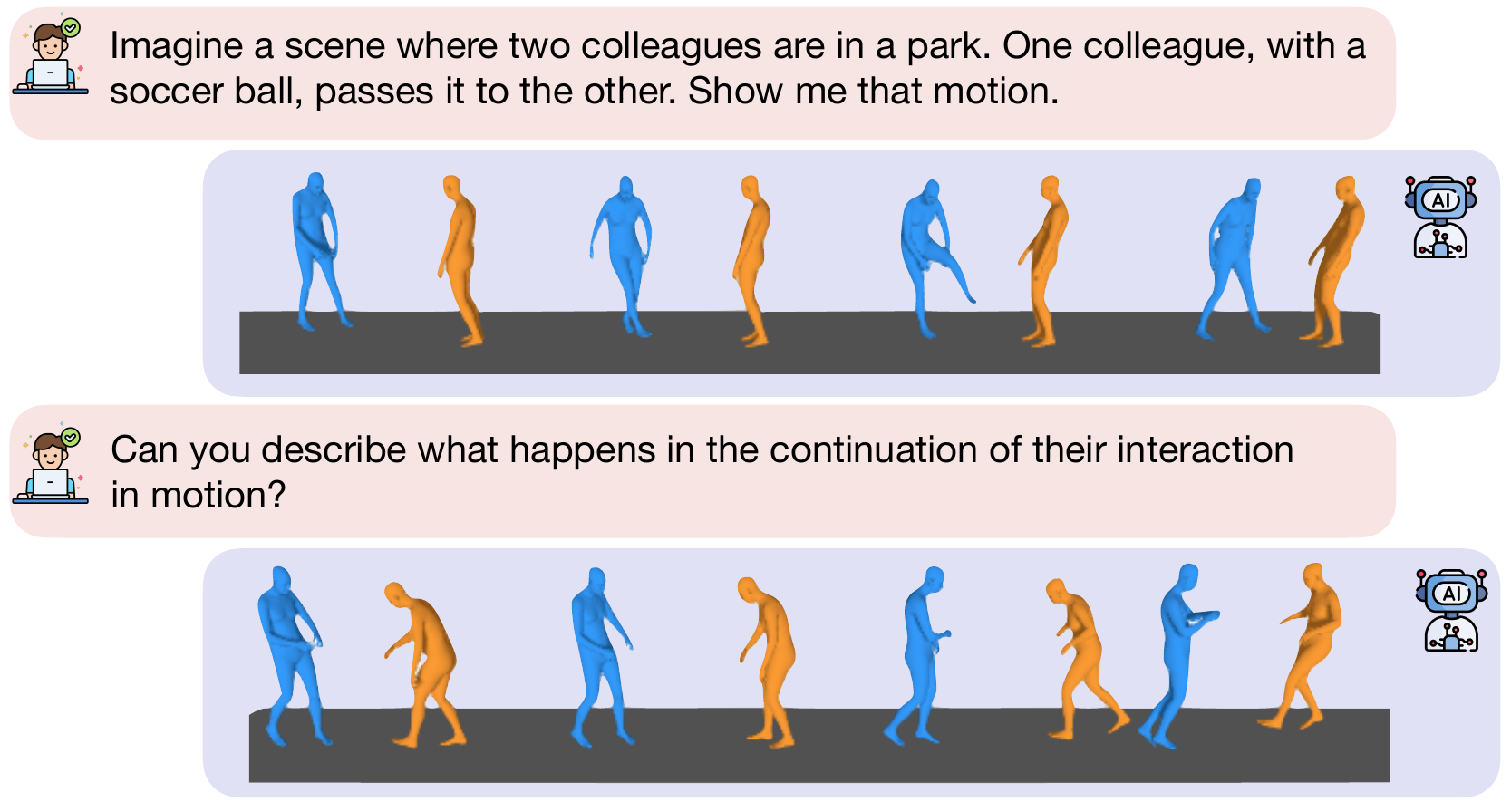}
    \caption{ A multi-turn interaction example where two people are playing soccer, illustrating the dataset’s detailed motion and conversational annotations.}
    \label{fig:short-a}
    \end{subfigure}
    \vspace{-1.5em}
    \caption{Statistics and data sample from \dataset. }
    \label{fig:syn_data}
\end{figure}

In this section, we present \dataset dataset, for modeling multi-turn interactive motion of multiple humans. 
Previous datasets \cite{liang2024intergen,xu2024inter} provide a textual description of the motions, lack sufficient diversity in instructions, and do not include multi-turn conversations. Since they are insufficient to enable a model to understand and generate complex interaction motions in multi-tern scenarios, we introduce \datasetbold: \textbf{Inter}active \textbf{M}uti-\textbf{{T}}urn \textbf{{M}}otion-\textbf{{T}}ext dataset. This dataset covers a variety of interactive motion scenarios with multi-turn conversations, diverse instructions, and spatiotemporally aligned motions between two individuals.
We enhance our dataset by generating diverse instructions from large language models and combining motion data from existing datasets with generative approaches to enable flexible text-to-motion modeling.
\begin{table}[]
    \centering
\resizebox{0.65\columnwidth}{!}{%
\begin{tabular}{l|c|c}
         \toprule
        Dataset & Ret. top-3& Div. \\
        \midrule
        Source dataset  & 0.870 &  0.997 \\        
        Generated by InterGEN & 0.645 & 0.953 \\
        \rowcolor{blue!10}
        \dataset (Ours) &  0.701 & 0.931 \\
        \bottomrule
    \end{tabular}%
    }
    \vspace{-0.5em}
    \caption{Comparison of generated motions on text-matching ability (top-3 retrieval precision), and motion diversity (Div.).}    
    \label{tab:dataset}
\end{table}

We begin with the human interaction motion and text datasets, Inter-X~\cite{xu2024inter} and InterHuman~\cite{liang2024intergen}, as the foundational resources for our dataset construction. 
To convert these datasets into instructional datasets, we first generate multi-turn instructions with motion captions using GPT-4o~\cite{openai244o}. 
We consider the instructional scenarios as various tasks with following text prompts, including motion editing (e.g., ``Make the left person more playful"), motion reasoning (e.g., ``What happened before/after this motion?"), and story generation (e.g., ``Let's create a story where two people are following this motion."). 
Detailed prompt templates and the complete data collection pipeline are presented in the supplementary materials. 
To guarantee high-quality caption generation, we guide the LLMs by providing action labels from the existing datasets alongside example captions, effectively constraining and enhancing the relevance and accuracy of the generated captions. 
Subsequently, we utilize a state-of-the-art diffusion-based text-to-motion model, InterGEN~\cite{liang2024intergen}, to synthesize interactive motions that align closely with these generated captions.

Our pipeline creates samples in two ways. First, starting with a dataset motion, we generate a caption and instruction and then use InterGEN~\cite{liang2024intergen} to synthesize a matching motion, yielding both the original and synthesized motions with the instruction. Alternatively, we generate two captions and instructions to synthesize two motions, producing samples entirely from synthesized motions. This method blends data-sourced and generative motions for reliable interactive motion modeling.
Overall, we collected 82K multi-turn conversations, including 96K synthesized and 56K real motions. Figure \ref{fig:syn_data} shows statistics and samples from our \dataset, where motion scenarios are classified using a large language model with motion captions.


To assess the quality and diversity of the generated motions and their alignment with texts, we evaluate our dataset using the text-motion matching score and diversity metric of our dataset, as shown in Table~\ref{tab:dataset}. 
Pre-trained retrieval models~\cite{petrovich23tmr} assess the alignment between motions and captions, with additional details in the supplementary material.
Our dataset achieves a top-3 retrieval precision of $0.701$ (the precision of the source dataset of the retrieval model is $0.870$), showing good alignment, which slightly surpasses the matching performance of the synthesized dataset created by the state-of-the-art motion generation method, InterGEN~\cite{liang2024intergen}. 
Additionally, our dataset exhibits robust diversity similar to the source dataset.
These results indicate that despite our multi-turn interactive motions and captions being synthetically generated, their quality closely approximates that of real-world datasets.

\begin{figure*}[!t]
    \centering
    \includegraphics[width=0.99\linewidth]{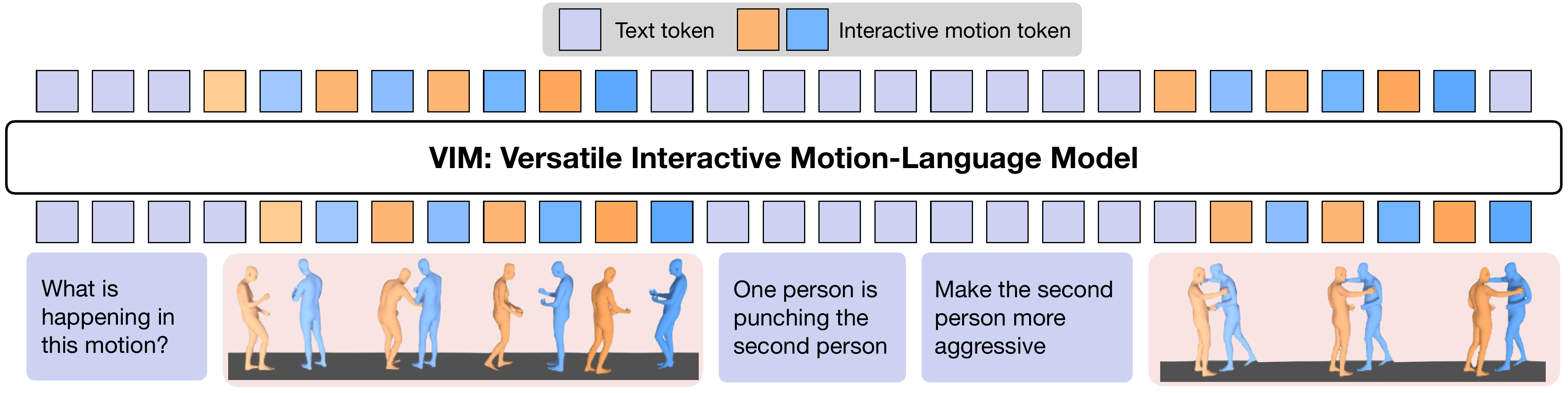}
    \vspace{-0.7em}
    \caption{
    An overview of \method, illustrating its versatile capability to flexibly process and generate interactive motions and texts in an auto-regressive manner. 
    We omit the motion tokenizer, which converts raw motion sequences into discrete motion tokens, for clarity. \method covers versatile motion tasks involving both motion and textual modalities across multiple conversational turns.  
    }
    \label{fig:method}
\end{figure*}

\section{VIM: Versatile Interactive Motion-Language Model}

In this section, we introduce \method, a versatile interactive motion-language model that processes multi-turn conversations with both language and two-person interactive motions as inputs and outputs.
First, we will explain our design choices for the model architectures, followed by a detailed description of the training methodologies. 

\subsection{Notations}
We denote an interactive motion from two individual $a$ and $b$ as $\{\mathbf{m}_a, \mathbf{m}_b\}$, following non-canonical representation in \cite{liang2024intergen} based on SMPL-X structure \cite{SMPL-X:2019} with $M$ as a motion length. 
At each motion time step $i$, the motion representation is defined as: $\mathbf{m}^i = [\mathbf{j}_g^p, \mathbf{j}_g^v, \mathbf{j}^r, \mathbf{c}^f]$, where $\mathbf{j}_g^p \in \mathbb{R}^{3N_j}$ is the global joint positions, $\mathbf{j}_g^v \in \mathbb{R}^{3N_j}$ is the global joint velocities, $\mathbf{j}^r \in \mathbb{R}^{6N_j}$ is 6D representation of local rotations with $N_j$ joints, and $\mathbf{c}^f \in \mathbb{R}^4$ is binary ground contact features. 
We train a motion-language model $p_\theta$ that jointly models text and motion data. The model processes the input (user instructions or context) and output (machine responses), effectively integrating both modalities.


\subsection{Architecture}
Our architecture for modeling and generating interactive motions consists of three primary components: motion tokenizer, large language model (LLM), and motion decoder. This design allows for the integration of both motion and text data within a unified framework. The overview of \method's architecture is shown in Figure~\ref{fig:method}. 

To enable the LLM to interpret interactive motions, we first tokenize the motion sequences. We utilize RQ-VAE~\cite{lee2022autoregressive} as a tokenizer to reduce the information loss during the quantization, similarly to the approach in~\cite{guo2024momask}.
\begin{wrapfigure}{r}{0.32\columnwidth}    
    \centering 
    \hspace{-5em}
    \includegraphics[width=1.25\linewidth]{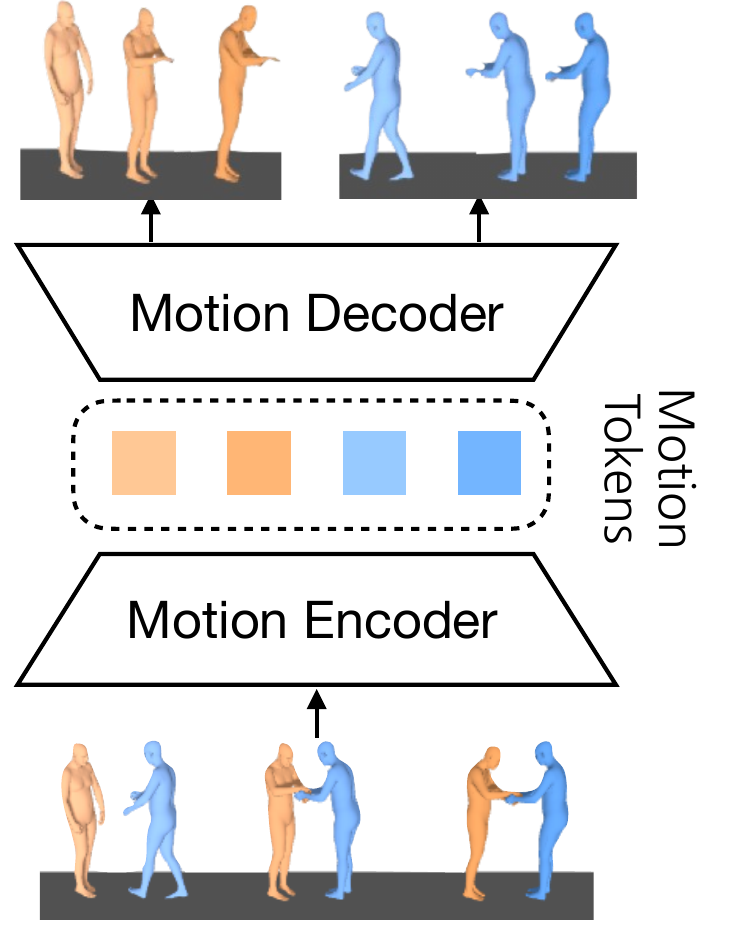} 
    \hspace{-5em}  
    \vspace{-0.5em}
    \caption{Tokenization of interactive motions.}
    \vspace{-1em}
\end{wrapfigure}
The motion encoder $\mathcal{E}_M$ applies 2D convolutions to motion features along the time axis, converting motion pairs $\{\mathbf{m}_a, \mathbf{m}_b\}$ into latent vectors $\{\mathbf{z}^{1:L}_a, \mathbf{z}^{1:L}_b\}$, $L=M/l$ with down-sample rate $l$. Each latent vector $\mathbf{z}^i$ is quantized into an ordered set of D discrete codes, $\mathcal{RQ} (\mathbf{z}^i; \mathcal{C},D) = (k^i_1, \cdots, k^i_D) \in [K]^D$, where $\mathcal{C}$ is the code book with $K = |\mathcal{C}|$, and $k^i_d$ is the code of $\mathbf{z}$ at timestep $i$ and depth $d$. 
These tokens, combined with special tokens indicating the start and end of motions, constitute the motion vocabulary. 
For text inputs, we utilize a standard text tokenizer compatible with the LLM.

Subsequently, the quantized tokens are provided to the LLM block, which serves as the central processing component. 
In this work, we initialize \method with the pretrained LLaMA-3.1-8B~\cite{dubey2024llama}. 
The motion vocabulary and text vocabulary of the LLM are integrated into a unified vocabulary, allowing the model to to efficiently process and generate both modalities.
Interactive motion is represented as $ X_m = \{k^{1;a}_{1:D}, k^{1;b}_{1:D
}, \cdots, k^{L;a}_{1:D
}, k^{L;b}_{1:D}\}$, where $X_m$ denotes the motion sequence encoded in the unified vocabulary, and $k^{i;a}_{1:D} \in [K]^D$ is the $i$-th token of motion $a$.

Finally, to visualize the generated motion tokens, we use the motion decoder of the RQ-VAE. The decoder projects the quantized features $\hat{\mathbf{z}}^i = \sum_{d=1}^D \mathbf{e}(k_d^i)$, converting them back into motion sequences. 


\subsection{Training}
We describe the training strategy in \method, to convert a large language model into an interactive motion-language model. 


\paragraph{Motion Tokenizer} 
The motion tokenizer consists of an encoder, decoder, and quantizer.
We followed the original objective functions from \cite{lee2022autoregressive}, minimizing the reconstruction loss, the codebook loss to align the encoder's outputs with the codebook, and the loss of commitment to ensure the consistency of the encoder.
After training the encoder and decoder, we freeze their parameters throughout the rest of the training stage. 

\paragraph{Pre-training for Cross-modal Motion-Text Alignment}
The goal of this stage is to enable the large language models (LLMs) to process and generate interaction motion tokens effectively. 
To achieve this, we continuously pre-train LLMs using paired interaction motion-text datasets, such as Inter-X~\cite{xu2024inter} and InterHuman~\cite{liang2024intergen}, across various tasks including motion-to-text, text-to-motion, motion prediction, and reaction generation.

For each task, we construct sequences $y$ that combine motion sequences with their corresponding captions and train with a next-token prediction objective $\mathcal{L} = -\log \sum^T p_\theta (y_i|y_{<i})$. 
To improve training efficiency, we employ LoRA adaptor~\cite{hu2022lora}, similar to \cite{ge2024seed}, and merge its parameters to the LLM backbone.
Furthermore, due to a limited number of interactive motion data, we also leverage a subset of single-person motion-text datasets from Motion-X~\cite{lin2024motion}. This additional single-person data offers prior knowledge of how the individual motions are described in language, enhancing the model's ability to align motions with textual descriptions.


\paragraph{Instruction-tuning with \dataset~Data}
In this stage, we aim to enhance the model to extend beyond understanding and generating single-turn interaction motions and focusing on handling diverse and complex instructions presented through \textbf{multi-turn} conversational scenarios. 
Similar to the pre-training stage, We adopt a next-token prediction training objective for training.
The instruction-tuning sequences are composed of user interactions paired with 
corresponding responses, integrating tokens from a unified vocabulary that covers texts, motions, or both modalities. 
We also leverage the \dataset dataset along with single-turn interaction data from existing motion datasets~\cite{xu2024inter, liang2024intergen}, formatted according to the instruction template of~\cite{jiang2023motiongpt}. 

\section{Experiments}
In this section, we evaluate the effectiveness of \method, particularly focusing on its capability to accurately understand and generate interactive motions in complex, multi-turn conversational scenarios involving both motions and text modalities. 
To extensively validate our approach, we compare \method against several specialized baseline methods, each explicitly designed for individual tasks. 
This allows us to understand the performance and versatility of \method. 
Additionally, we investigate the contribution and effectiveness of our proposed dataset, \dataset, showing how it enhances \method's ability to process and generate interactive motion and texts. 
We also provide qualitative video results generated by \method in the supplementary material. 


\subsection{Evaluation Tasks and Baselines}

\paragraph{Motion Reasoning}
We introduce a motion reasoning task to validate the model's ability to comprehend interactive motions and text queries.
Motion reasoning involves predicting past or future events, or reasoning about current motions, based on prior conversational data. 
This task requires the model to understand the context of the conversation, interpret how the given interactive motion fits within that context, and adjust its reasoning accordingly.
We utilize LLMs-based evaluator, specifically GPT-4o~\cite{openai244o}, to assess the content alignment, naturalness, and logical coherence of the generated textual responses. Content alignment evaluates how accurately the text reflects the given interactive motions, logical coherence checks the consistency and reasoning accuracy of inferences made about past or future events, and naturalness evaluates the fluency of generated texts, with rating each metric on a 10-point scale. 
Additionally, we employ linguistic metrics, such as METEOR \cite{banarjee2005}, and MAUVE \cite{pillutla2021mauve} to quantitatively evaluate relevance and fluency against 2002 labeled samples from the \dataset test set. 
We present the results on motion reasoning in \S\ref{subsec:motion_reasoning}.

\paragraph{Motion Editing} 
In the motion editing task, the model modifies the given motion based on a person’s persona or scenario, e.g., emotions or relationship dynamics, which adds complexity as changes in one individual affect the other's motion.
Unlike single-person motion editing~\cite{athanasiou2024motionfix, goel24meo}, the task that edits interactive motions should consider preserving contextual coherence and social dynamics.
We evaluated the methods on 1445 samples from the \dataset test set. In a within-subject user study (following \cite{goel24meo}), 30 participants each rated five samples (from 30 randomly selected tests) on content similarity, instruction alignment, and motion quality using a 5-point Likert scale.
Content similarity evaluates whether the edited motion preserves the original meaning of the source motion, while instruction alignment assesses how accurately the edited motion follows the given command.
Participants compared our method against four baselines by reviewing randomly shuffled motion outputs. 
Additionally, we measured performance using data-driven metrics, Frechet Inception Distance (FID), and mean per joint position error (MPJPE), against the labeled motions in the \dataset test set, following \cite{goel24meo}.
The results are detailed in \S\ref{subsec:motion_editing}.

\paragraph{Traditional Motion Relevant Tasks}
We further evaluated our method on three traditional interactive motion tasks: motion-to-text, text-to-motion, and reaction generation,
using the combined test sets from InterHuman \cite{liang2024intergen} and Inter-X \cite{xu2024inter}. Text-motion matching is assessed via top-3 retrieval precision (batch size 32) in the retrieval models' feature space \cite{petrovich23tmr}. Motion quality is measured by the Frechet Inception Distance (FID) and the accuracy of reaction motions is measured by mean per joint position error (MPJPE) in meters.
Detailed results are in \S\ref{subsec:motion_related}.

\paragraph{Baselines}
Since our interactive multi-turn scenarios and tasks, including interactive motion reasoning and motion editing, are novel, there is no exact comparison method. We compare our method against reasonable baselines that handle both motion and texts as input and output. 
\begin{itemize}
    \item \textbf{Two-stage approach}. We leverage off-the-shelf LLMs and motion-to-text methods. For the motion reasoning task, we convert motions to text via the state-of-the-art motion-to-text model, TM2T \cite{guo2022tm2t} and then apply large language models (GPT-4o~\cite{openai244o}, LLaMA-3.1-8B~\cite{dubey2024llama}). For the motion editing task, we first convert the given motions into text descriptions using TM2T, and we concatenate the motion description with editing command texts. We then put the texts to InterGEN~\cite{liang2024intergen} to generate modified motions. 
    \item \textbf{Extending unified single-human motion model}. We adopt a single-human motion-language model, MotionGPT \cite{jiang2023motiongpt}, for interactive motions. We consider three variations: (1) $\text{MotionGPT}^*$: a modified MotionGPT fine-tuned on interactive motion data; (2) $\text{MotionGPT}^*_I$: 
$\text{MotionGPT}^*$ enhanced with \dataset dataset; (3) \method~w/o \dataset: our method fine-tuned with instruction templates, but without \dataset~data.
    \item \textbf{Motion generation baselines.} For traditional motion tasks, we benchmarked against interactive motion generation methods, ComMDM~\cite{shafir2024human} and InterGEN~\cite{liang2024intergen}, and a single human motion generation model, MoMask$^*$~\cite{guo2024momask}.
\end{itemize}

\subsection{Motion Reasoning}
\label{subsec:motion_reasoning}

\begin{table}[!t]
\centering
\resizebox{\columnwidth}{!}{%
\begin{tabular}{l|c c c|c c }
    \toprule
     \multicolumn{1}{c|}{Methods} & \multicolumn{3}{c|}{LLM-Assisted} & \multicolumn{2}{c}{Linguistic Metrics} \\
     & Coh. $\uparrow$ & Align. $\uparrow$ & Nat. $\uparrow$  &  METEOR & MAUVE \\
     \midrule
     \textcolor{gray}{\textit{two-stage approach}} & & & & & \\
    TM2T + LLaMA-3.1 & 3.852 & 3.050  & 6.348 & 0.226 & 0.009 \\
    TM2T + GPT-4o & \underline{4.266} & \underline{3.455} & \underline{6.790} &  0.227 & 0.019\\
    \midrule
    \textcolor{gray}{\textit{unified approach}} & & & &\\
    $\text{MotionGPT}^*$ &  1.855 & 1.303 & 3.574 & 0.096 & 0.005\\
    $\text{MotionGPT}^*_I$ & 3.690 &3.160 &  5.291 & \underline{0.218} & \underline{0.417} \\
    VIM w/o \dataset & 2.770 & 2.141 & 4.968 &  0.145 &  0.004 \\
    \rowcolor{blue!10}    
    \textbf{VIM (Ours)} & \textbf{5.252} & \textbf{4.511}  & \textbf{6.981} & \textbf{0.260} & \textbf{0.794}  \\
    \bottomrule
\end{tabular} %
}
\vspace{-0.5em}
\caption{
Evaluation on Motion Reasoning task with \dataset test set. 
Coh., Align., and Nat. denote logical coherence, content alignment, and naturalness, respectively. 
\textbf{Bold} indicates best performance and \underline{underline} denotes the second best performance.
}
\label{tab:reasoning}
\end{table}
\begin{figure*}[!t]
    \centering
    \includegraphics[width=0.82\linewidth]{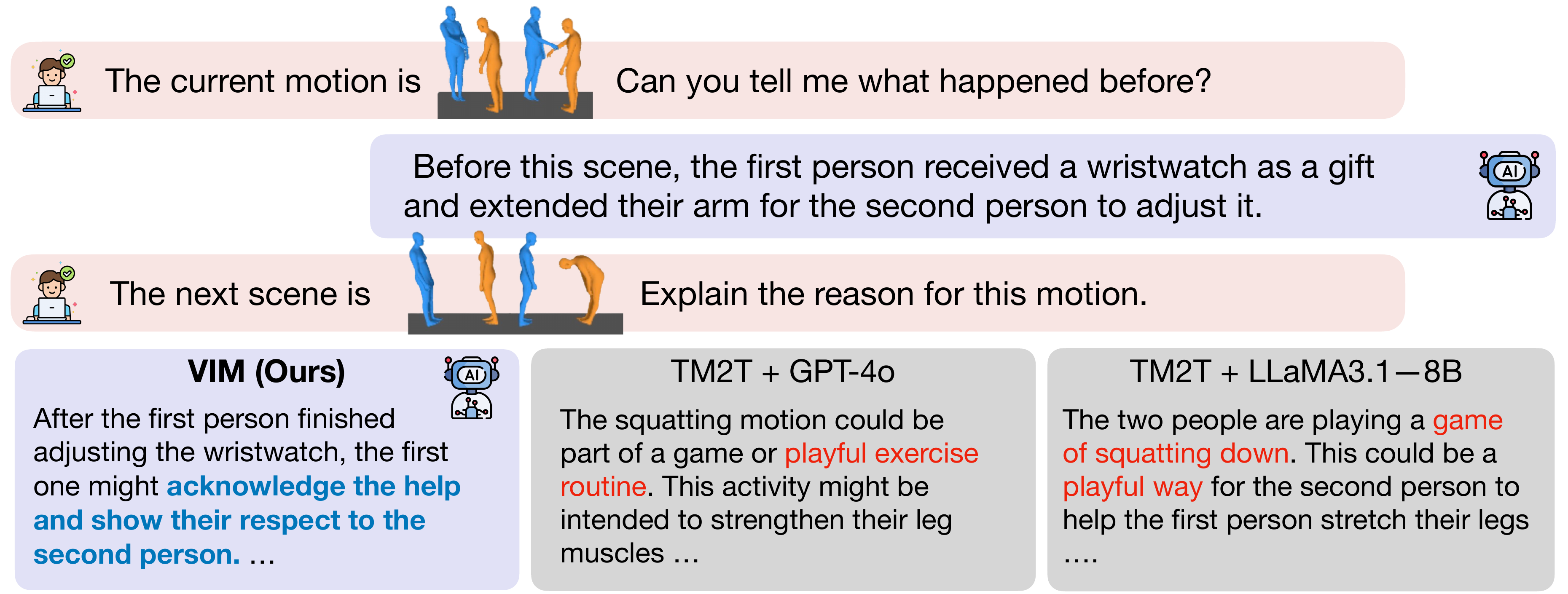}
    \vspace{-0.5em}
    \caption{Generated samples for interactive motion reasoning task. This example shows how \method~explains behaviors and their motivations, demonstrating a deeper understanding of scenarios by incorporating context from prior interactions.
    }    
    \label{fig:rea}
\end{figure*}
In the motion reasoning task, conversations about two interactive motions are examined to assess the model's ability to deduce past or future events and comprehend the motivations driving the motions.
The experimental results in Table~\ref{tab:reasoning} demonstrate that our unified model, \method, outperforms baselines across all LLM-assisted and linguistic metrics. Specifically, \method~achieves improvements with performance increases exceeding 1.9 points in logical coherence, 1.1 points in content alignment, and nearly 0.2 points in naturalness compared to the best two-stage model. 

The improved performance of our unified model, \method, over two-stage approaches, appears to result from two key factors: error accumulation and interpretation ambiguity.
Two-stage models can carry over errors if the motion captioning step is inaccurate, undermining content alignment and coherence. In contrast, our unified architecture integrates motion encoding and reasoning in a single framework, minimizing error propagation. Moreover, a single caption may not fully capture multiple interpretations of the same motion, compromising context accuracy in two-stage setups. Our unified approach, however, accounts for these varied interpretations to generate more contextually precise outputs.
Figure~\ref{fig:rea} shows its ability to dynamically adjust interpretations and responses by incorporating context from previous conversations.


\subsection{Motion Editing}
\label{subsec:motion_editing}


\begin{figure}[t]
\centering
\includegraphics[width=0.95\linewidth]{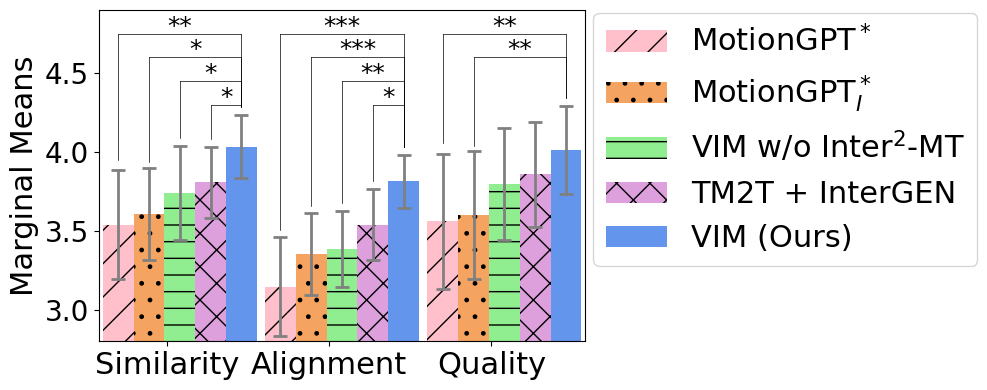}
\vspace{-0.5em}
\caption{User subject study results for motion editing. We plotted the difference only in a post hoc pairwise comparison of the proposed method. * as $0.01<p<0.05$, ** as $p<0.01$, and *** as $p<0.001$. The error bars represent 95\% confidence intervals.}
\label{fig:user}
\end{figure}



We aimed to validate the hypothesis that users perceive interactive motions edited by our proposed method as more content-consistent, better aligned with instructions, and of higher overall quality.
To investigate this, we conducted a user study and analyzed the results using repeated-measures multivariate analysis of variance (MANOVA). The analysis revealed significant effects of the method on user perception across all evaluated dimensions;
$F(4) = 4.591, p = 0.002, \eta^2 = 0.137$ for content similarity, $F(4) = 7.134, p = 0.000, \eta^2 = 0.197 $ for instruction alignment, and $F(4) = 4.781, p = 0.001, \eta^2 = 0.142 $ for motion quality, with all $\alpha = 0.05$. The estimated marginal mean of the rated score is reported in Figure \ref{fig:user}. The results show that the proposed method had better alignment, quality, and consistency of instruction in other baselines with significant differences. 


\begin{table}[!t]
\centering
\resizebox{0.65\columnwidth}{!}{%
\begin{tabular}[t]{l|c c}
        \toprule
         Methods & FID $\downarrow$  & MPJPE $\downarrow$ \\
         \midrule
         \textcolor{gray}{\textit{two-stage approach}} & & \\
        TM2T +  InterGEN &  0.110 & \underline{0.811} \\
        \midrule
        \textcolor{gray}{\textit{unified approach}} & & \\
        $\text{MotionGPT}^*$&  0.251 & 4.002 \\ 
        $\text{MotionGPT}^*_I$&  0.161 & 3.982 \\ 
        VIM w/o \dataset & \underline{0.080} & 0.908  \\ 
        \rowcolor{blue!10}        
        \textbf{VIM (Ours) } &  \textbf{0.064} & \textbf{0.758} \\ 
        \bottomrule
    \end{tabular} %
}
\vspace{-0.5em}
\caption{Quantitative results in motion editing task. }
    \label{table:edit}
\end{table}
\begin{figure*}[!h]
    \centering
    \includegraphics[width=0.98\linewidth]{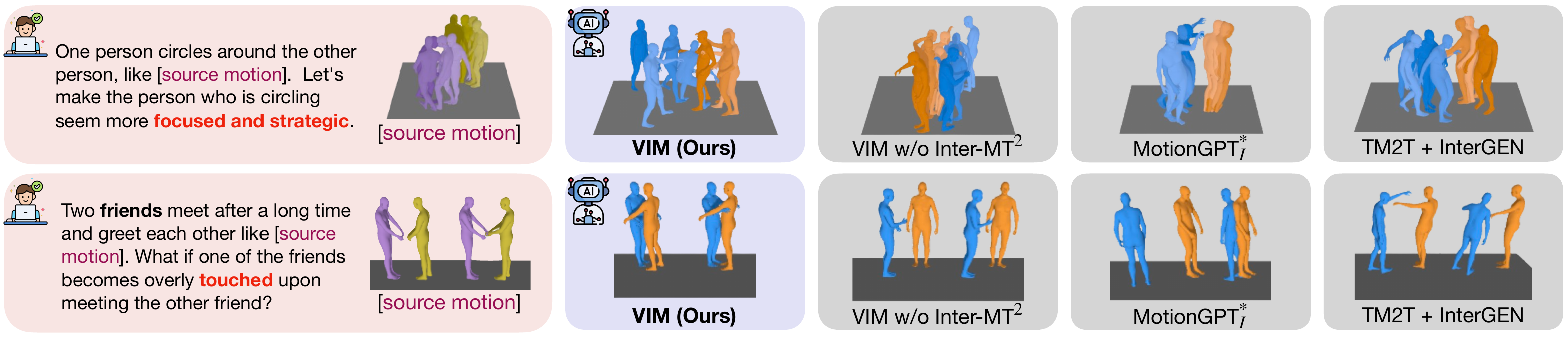}
    \vspace{-0.6em}
    \caption{Generated samples for interactive motion editing. The proposed method excels in capturing nuances, outperforming alternatives in content similarity and instruction alignment.}
    \label{fig:edits}
\end{figure*}
During post-hoc pairwise comparisons, \method significantly outperforms the two-stage model (TM2T~\cite{guo2022tm2t} with InterGEN~\cite{liang2024intergen}) in terms of content similarity ($p = 0.017$) and instruction alignment ($p = 0.010$). 
The two-stage model had lower content similarity due to motion-to-text conversion errors causing unintended motions. It also struggled with instruction alignment since InterGEN was trained to generate motions from textual descriptions, limiting its adaptability. In contrast, our unified framework avoids error accumulation and, trained on diverse instructions, demonstrates superior reasoning and adaptability for accurate motion editing and generation.
Compared to \method~w/o \dataset, our model significantly improves content similarity ($p=0.010$) and instruction alignment ($p=0.001$), suggesting that excluding \dataset data hinders motion control. It also outperforms $\text{MotionGPT}^*_I$ in all metrics, indicating that the baseline’s VQ-based tokenizer struggles to capture precise relative joint positions in two-person motion. Further ablation studies on the motion tokenizer are provided in the supplementary materials.
Quantitative evaluations using data-driven metrics, specifically FID and MPJPE (Table~\ref{table:edit}), further confirmed the superiority of our method over baseline methods, consistent with user study results. Examples of generated edited motions are illustrated in Figure~\ref{fig:edits}.

\subsection{Traditional Motion Related Tasks}
\label{subsec:motion_related}

\begin{table}[!t]
\centering
\resizebox{0.5\textwidth}{!}{%
\begin{tabular}{c|c|cc|cc} 
    \toprule
     \multirow{2}{*}{Methods}  & M2T &  \multicolumn{2}{c|}{T2M} &  \multicolumn{2}{c}{Reaction Gen.} \\
     & R Top3  $\uparrow$ &  R Top3 $\uparrow$ & FID $\downarrow$ & MPJPE $\downarrow$ & FID $\downarrow$ \\ 
     \midrule
     Real &  0.867 & 0.869 & 0.00 & - & 0.00\\
     \midrule
     \textcolor{gray}{\textit{task-specific approach}}& & & &  \\ 
     $\text{TM2T}^*$ & 0.696 & 0.534 & 0.300 & - & - \\
     $\text{MoMask}^*$ & - & \underline{0.612} & \underline{0.066} & 1.602 & 0.112 \\
     ComMDM & - & 0.251 & 0.304 &  - & - \\
     InterGEN & - & \textbf{0.645} & 0.078 & - & - \\
     \midrule
     \textcolor{gray}{\textit{unified approach}}& & & &  \\ 
    $\text{MotionGPT}^*$ & 0.494 & 0.328 &  0.123 & 3.444 & 0.355 \\
    $\text{MotionGPT}^*_I$ & 0.503 & 0.331& 0.118 & 1.436 & 0.380\\
    VIM w/o \dataset &  \underline{0.894} & 0.561 & 0.082 &  \underline{0.984} & \underline{0.031} \\
    \rowcolor{blue!10}   
    \textbf{VIM (Ours)} & \textbf{0.901} & 0.568 & \textbf{0.059} & \textbf{0.691} & \textbf{0.019}\\
    \bottomrule
\end{tabular}%
    }
    \vspace{-0.6em}
\caption{Comparisons for three motion-related tasks on Inter-X and InterHuman datasets. M2T denotes motion-to-text, T2M for text-to-motion, and Reaction Gen. for reaction generation.}    
\label{tab:motion_tasks}
\end{table}

In this section, we conduct comparison experiments on existing motion-relevant tasks, such as motion-to-text (M2T), text-to-motion (T2M), and reaction generation. The detailed results are in 
Table~\ref{tab:motion_tasks}.
The first row (``Real'') shows retrieval accuracy, and FID scores from the dataset labels. Note that both \method~w/o \dataset and MotionGPT$^*$ were trained on all of these tasks for fair comparison.
The results confirm that incorporating \dataset~dataset enhances the model's performance in traditional motion tasks, by comparing with \method~w/o \dataset. 

For M2T, Top-3 retrieval accuracy improved from 0.894 (\method w/o \dataset) to 0.901 (\method). For T2M, it rose from 0.561 to 0.568, with FID dropping from 0.082 to 0.059, indicating better motion generation. For reaction generation, MPJPE decreased from 0.984 to 0.691 and FID from 0.031 to 0.019, highlighting the benefits of multi-turn datasets for motion comprehension and generation. We believe that \dataset~dataset provides diverse, context-rich examples, helping the model learn more nuanced relationships between text and motion.

In addition, we compared \method against task-specific methods, each optimized individually per each task. 
Note that the methods marked with an asterisk (*) were originally designed for single-motion tasks and were trained on interactive motion data for our evaluation. \method outperforms these specialized models in motion-to-text (M2T) and reaction generation tasks, achieving higher retrieval precision accuracy and lower MPJPE and FID scores. In the text-to-motion (T2M) task, \method achieves comparable performance against the state-of-the-art task-specific models, including InterGEN and MoMask$^*$, highlighting the capability of \method in generating high-quality interactive motions.


\subsection{Discussion: Generating Multi-Human Motions}
\begin{figure}[!h]
    \centering
    
    \vspace{-0.6em}
    \includegraphics[width=0.99\linewidth]{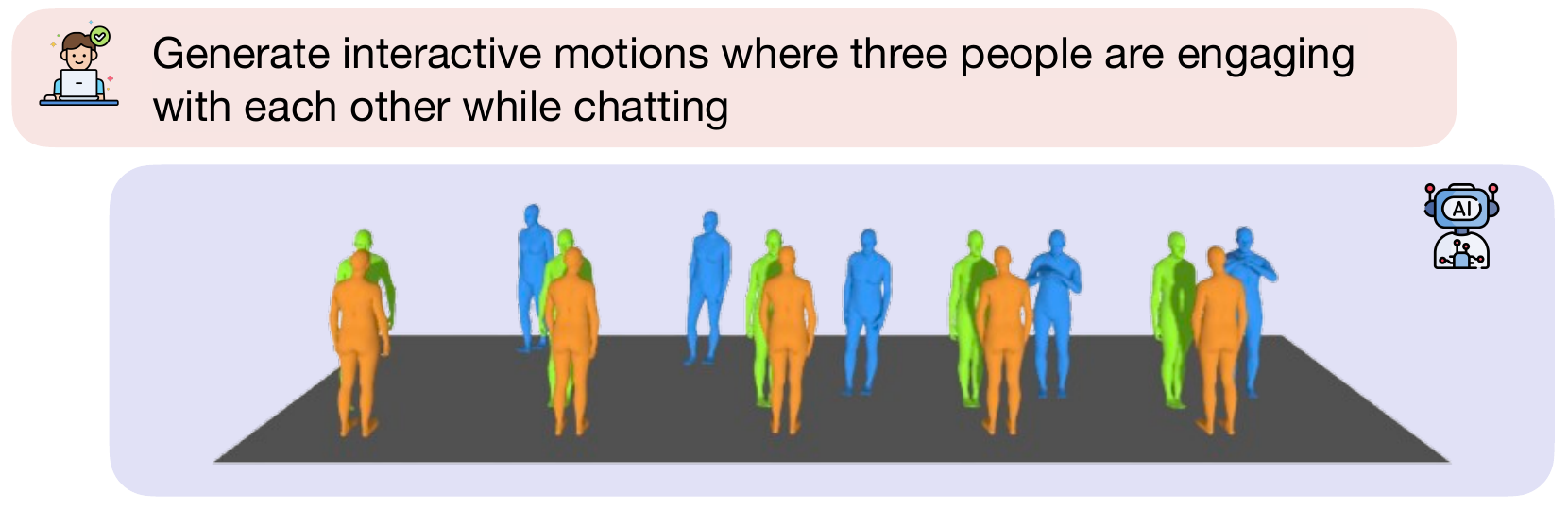}
    \caption{Expanding \method to generate multiple human motions. For clarity, we simplified the incremental process, where \method first generates two-person motion and adds a third-person to them.}
    \vspace{-0.6em}
    \label{fig:multi_main}
\end{figure}

Interestingly, the versatility of \method allows it to generalize beyond two-person interactions without explicit fine-tuning on multi-person interactive data. 
Specifically, users can incrementally generate motions: first generating a two-person motion, then adding a third person's motion conditioned on the existing context. By leveraging prior turns, \method seamlessly integrates the new figure’s movements, as shown in Figure~\ref{fig:multi_main}. 
Since our method is agnostic to the number of people, it can readily extend to groups or crowds, provided interactive multi-person data is available.
\section{Conclusion}
In this paper, we introduced \method, a versatile motion-language model designed to understand, generate, and reason about interactive motions. We presented the detailed architecture and training strategy of our unified framework, which integrates large language models with interactive motion modality. 
To further enhance the model's reasoning capabilities and applicability, we presented a specialized dataset, \dataset, which incorporates a variety of reasoning tasks set within multi-turn conversations centered on interactive motions. Our comprehensive experiments demonstrated that \method successfully handles instruction-following, motion editing, and motion reasoning tasks, highlighting its capability to effectively interpret and generate contextually accurate interactive motions. 

{
    \small
    \bibliographystyle{ieeenat_fullname}
    \bibliography{main}
}

\clearpage
\onecolumn 

\renewcommand{\thesection}{\Alph{section}} 
\renewcommand{\thefigure}{\thesection.\arabic{figure}}
\renewcommand{\thetable}{\thesection.\arabic{table}}
\setcounter{figure}{0}
\setcounter{table}{0}
\setcounter{section}{0}
\begin{center}
  {\Large \bfseries Appendix\par}
\end{center}

This appendix provides a comprehensive set of supplementary materials that reinforce the main findings of the research. 
The appendix begins with motion representation and motion token representation (Sec.\ref{sec:1}), followed by ablation studies on the pretraining method (Sec.\ref{sec:2}), along with ablation studies on the motion tokenizer (Sec.\ref{sec:3}), the demonstration for expansion to multiple human ($\geq 3$) motion generation (Sec.\ref{sec:4}), and illustrations of the data collection pipeline (Sec. \ref{sec:5}). 
More detailed results for traditional motion-related tasks are presented (Sec.\ref{sec:6}), limitations (Sec.\ref{sec:7}), and implementation details for the proposed methods (Sec.\ref{sec:8}) and baselines models trained for interactive motions (Sec.\ref{sec:9}). Task explanations cover motion editing and reasoning (Sec.\ref{sec:10}), with implementation details of two-stage baselines (Sec.\ref{sec:11}). The evaluation metrics for traditional
motion-related tasks are presented in Sec. \ref{sec:12}. Further sections include templates for pre-training and instruction tuning (Sec.\ref{sec:13}), data visualization and statistics (Sec.\ref{sec:14}, Sec.\ref{sec:15}), qualitative results (Sec.\ref{sec:16}), user study protocols (Sec.\ref{sec:17}), and prompts for data collection and LLM-assisted evaluation (Sec.\ref{sec:18}, Sec.~\ref{sec:19}).

\section{Motion Representation and Motion Token Representation}\label{sec:1}
For two persons $a$ and $b$, 
we denote the interactive motion as $\{\mathbf{m}_a, \mathbf{m}_b\}$, following non-canonical representation from \cite{liang2024intergen}. Each timestep of the motion $\mathbf{m}^i = [\mathbf{j}_g^p, \mathbf{j}_g^v, \mathbf{j}^r, \mathbf{c}^f]$ is composed of global joint positions $\mathbf{j}_g^p \in \mathbb{R}^{3N_j}$, global joint velocities $\mathbf{j}_g^v \in \mathbb{R}^{3N_j}$, 6D representation of local rotations $\mathbf{j}^r \in \mathbb{R}^{6N_j}$, with the number of joints $N_j$, and binary ground contact features $\mathbf{c}^f \in \mathbb{R}^4$. This non-canonical representation is applied for both interactive motions and single-person motions. 
All the motions are represented in an SMPL-X~\cite{SMPL-X:2019} format.

Motion tokenizer encodes the interactive motion into discrete residual tokens in depth $D$, based on latent vector $\mathbf{z}$. 
\begin{equation}
    \mathcal{RQ} (\mathbf{z}^i; \mathcal{C},D) = (k^i_1, \cdots, k^i_D) \in [K]^D
\end{equation}
where $\mathcal{C}$ is the codebook, $K = |C|$, $D$ is a depth, and $k^i_d$ is code of $\mathbf{z}$ at timestep $i$ with depth $d$. 

The interactive motion token sequence is represented as $ X_m = \{k^{1;a}_{1:D}, k^{1;b}_{1:D
}, \cdots, k^{L;a}_{1:D
}, k^{L;b}_{1:D}\}$, where $X_m$ is a sequence of motion represented in unified vocabulary and  $k^{i;a}_{1:D} \in [K]^D$ is the $i$-th token of motion $a$. In particular, the motion token is represented as below:
\begin{align*}
X_m =  \{& \texttt{<motion\_token\_start>}, \\
&\texttt{<motion\_token\_a\_start>},
& k^{1;a}_1, \cdots, k^{1;a}_D, \quad 
& \texttt{<motion\_token\_a\_end>}, \\
&\texttt{<motion\_token\_b\_start>}, 
& k^{1;b}_1, \cdots, k^{1;b}_D, \quad 
& \texttt{<motion\_token\_b\_end>}, \\
& \cdots \\
&\texttt{<motion\_token\_a\_start>}, 
& k^{L;a}_1, \cdots, k^{L;a}_D, \quad 
& \texttt{<motion\_token\_a\_end>}, \\
&\texttt{<motion\_token\_b\_start>}, 
& k^{L;b}_1, \cdots, k^{:;b}_D, \quad 
& \texttt{<motion\_token\_b\_end>}, \\
& \texttt{<motion\_token\_end}\}
\end{align*}
where \texttt{<motion\_token\_start>}, \texttt{<motion\_token\_a\_start>}, \texttt{<motion\_token\_b\_start>},  \texttt{<motion\_token\_a\_end>}, \texttt{<motion\_token\_b\_end>}, and  \texttt{<motion\_token\_end>}
is a special token added to the unified vocabulary. 
For modeling single-motion in pre-training we omitted the input string about \texttt{motion\_token\_b}.

\section{Ablation Studies on Pretraining Method}\label{sec:2}

\begin{table}[!h]
\centering
\caption{Ablation studies in pertaining stage for three motion-related tasks on InterX and Interhuman dataset.}
\begin{tabular}{ccc|c|cc|cc} 
    \toprule
     \multirow{2}{*}{Methods} & \multirow{2}{*}{Data} & Trainable & M2T &  \multicolumn{2}{c|}{T2M} &  \multicolumn{2}{c}{Reaction Gen.} \\
     & & Params &R Top3  $\uparrow$ &  R Top3 $\uparrow$ & FID $\downarrow$ & MPJPE $\downarrow$ & FID $\downarrow$ \\ 
     \midrule
     Real & - & - & 0.867 & 0.869 & 0.00 & - & 0.00\\
     \midrule
    $\text{MotionGPT}^*$ & InterX+H & 248M & 0.518 & 0.280 & 0.178 & 1.338 & 0.364 \\
    VIM-VQ & InterX+H & 726M &  0.709 & \textbf{0.511} & 0.181 & 1.750 & 0.181\\
    VIM (Ours) &  InterX+H &  726M & 0.721 & 0.427 & \textbf{0.161} & 1.494 & 0.157\\ 
    \textbf{VIM} (Ours) & InterX+H + MotionX & 726M& \textbf{0.729} & 0.464 & 0.172 & \textbf{1.236} & \textbf{0.131}  \\
    \bottomrule
\end{tabular}%
\label{tab:supp_motion_tasks}
\end{table}

We conducted ablation studies on the pertaining method. All the baselines are pre-trained models, not including the fine-tuning stage.
To evaluate the effectiveness of our pretraining approach, we conducted ablation studies comparing different methods on three motion-related tasks: Motion-to-Text (M2T), Text-to-Motion (T2M), and Reaction Generation. As shown in Table \ref{tab:supp_motion_tasks}, we compared our proposed method, VIM, against MotionGPT$^*$ and VIM-VQ, using the InterX~\cite{xu2024inter} and Interhuman (H) datasets~\cite{liang2024intergen}. MotionGPT$^*$ serves as a baseline with 248M trainable parameters, achieving a retrieval Top3 score of 0.518 in M2T and 0.280 in T2M, with corresponding FID scores of 0.178 and 1.338 for T2M and Reaction Generation, respectively. VIM-VQ, with 726M parameters, improves the M2T retrieval Top3 to 0.709 and T2M retrieval Top3 to 0.511, while maintaining competitive FID scores.

Our method, VIM, further enhances performance by achieving a retrieval Top3 of 0.721 in M2T and reducing the T2M FID to 0.161, alongside an MPJPE of 1.494 and FID of 0.157 in Reaction Generation. Notably, when incorporating the additional MotionX~\cite{lin2024motion} dataset, VIM achieves the highest M2T R Top3 of 0.729 and the lowest FID scores of 0.172 in T2M and 0.131 in Reaction Generation, demonstrating the substantial benefits of our comprehensive pretraining strategy. These results indicate that our approach not only outperforms existing models in generating accurate and high-quality motions but also effectively leverages additional data to enhance interactive motion understanding and generation. The ablation studies highlight the critical role of our pretraining methodology and the integration of diverse datasets in achieving superior performance across multiple interactive tasks.

\section{Ablation Studies on Motion Tokenizer}\label{sec:3}

\begin{table*}[!h]
\centering
\caption{Ablation Studies on motion tokenizer base model. We compared VQ-VAE-based tokenizer and the RQ-VAE-based model. }
\begin{tabular}{c|ccc|cc|c|cc|cc} 
    \toprule
     \multirow{2}{*}{Methods} &
     \multicolumn{3}{c|}{Reasoning} & \multicolumn{2}{c|}{Editing} & 
     M2T &  \multicolumn{2}{c|}{T2M} &  \multicolumn{2}{c}{Reaction Gen.} \\
     & Coh. $\uparrow$& Align. $\uparrow$ & Nat.$\uparrow$ & MPJPE $\downarrow$ & FID $\downarrow$ & R Top3  $\uparrow$ &  R Top3 $\uparrow$ & FID $\downarrow$ & MPJPE $\downarrow$ & FID $\downarrow$ \\ 
     \midrule
    VIM-VQ & 5.004 & 4.256 &6.915 & 0.892 & 0.128 & 0.861 & \textbf{0.601} & 0.101 & 1.109 & 0.055 \\
    \textbf{VIM (Ours)} & \textbf{5.252} & \textbf{4.511}  & \textbf{6.981} & \textbf{0.758} &\textbf{0.064}& \textbf{0.901} & 0.568 & \textbf{0.059} & \textbf{0.691} & \textbf{0.019}\\
    \bottomrule
\end{tabular}%
\label{tab:abla}
\end{table*}

We conducted ablation studies comparing the VQ-VAE-based model with our RQ-VAE-based approach, as shown in Table \ref{tab:abla}. The RQ-VAE-based motion tokenizer outperformed the VQ-VAE model in motion reasoning tasks, achieving higher scores in coherence, alignment, and naturalness. This improvement is attributed to reduced information loss, allowing our model to capture finer motion details while also enhancing its motion-to-text retrieval precision. 

For generation and editing tasks, the VQ-VAE model achieved slightly better text-to-motion retrieval accuracy but performed worse in FID and MPJPE across editing, reaction generation, and T2M tasks, indicating degraded motion quality and less precise motion details. 
In contrast, our approach reduced MPJPE by 0.055 for reaction generation, preserving joint dynamics and producing more realistic and natural motions. VQ-VAE’s limitations are especially problematic for modeling interactive motions, where precise relative positioning is crucial, making its information loss and reconstruction quality more evident.

\begin{figure}[!h]
    \centering
    \includegraphics[width=0.9\linewidth]{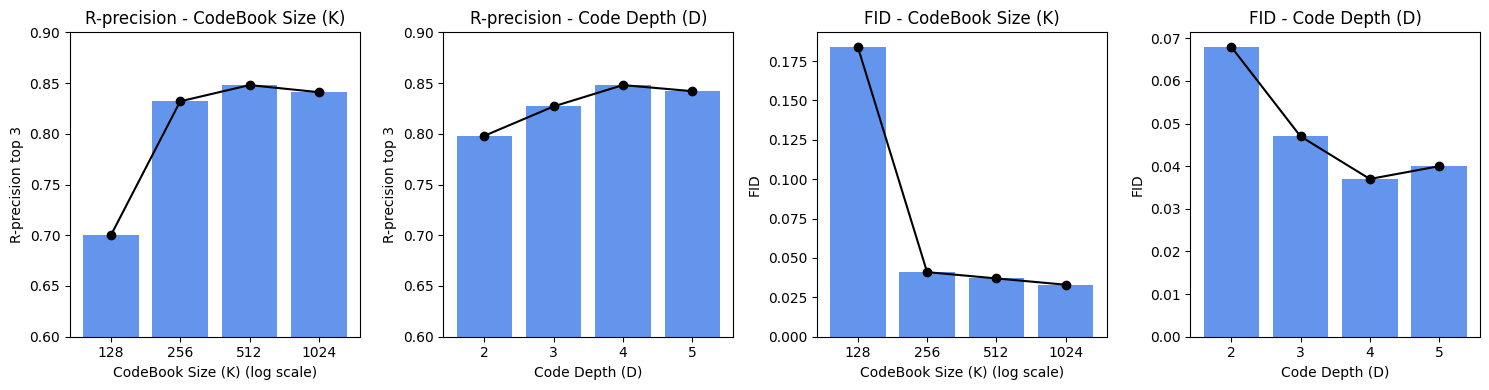}
    \caption{Ablation Studies on codebook size and depth. We measured Top-3 retrieval-precision accuracy and FiD on the reconstructed motion. }
    \label{fig:enter-label}
\end{figure}
We analyzed the effect of varying codebook size (128, 256, 512, 1024) and code depth (2, 3, 4, 5) on both reconstructed motion quality (measured by FID) and retrieval precision at top-3, and observed that increasing the codebook size from 128 to 512 reduced FID while simultaneously improving retrieval precision, indicating richer and more accurate motion representations, whereas moving to 1024 yielded diminishing returns at higher computational cost. Likewise, increasing the code depth from 2 to 4 provided better reconstruction quality and retrieval performance by allowing the model to capture more complex motion patterns, but further increasing the depth (e.g., to 5) showed marginal or even negative gains. Consequently, we selected 512 codes and a depth of 4 as the best trade-off between quality, retrieval accuracy, and efficiency.

\section{Expansion to Multi-Human Motion ($\geq 3$) Generation}\label{sec:4}
\begin{figure}[!h]
    \centering
    \includegraphics[width=0.7\linewidth]{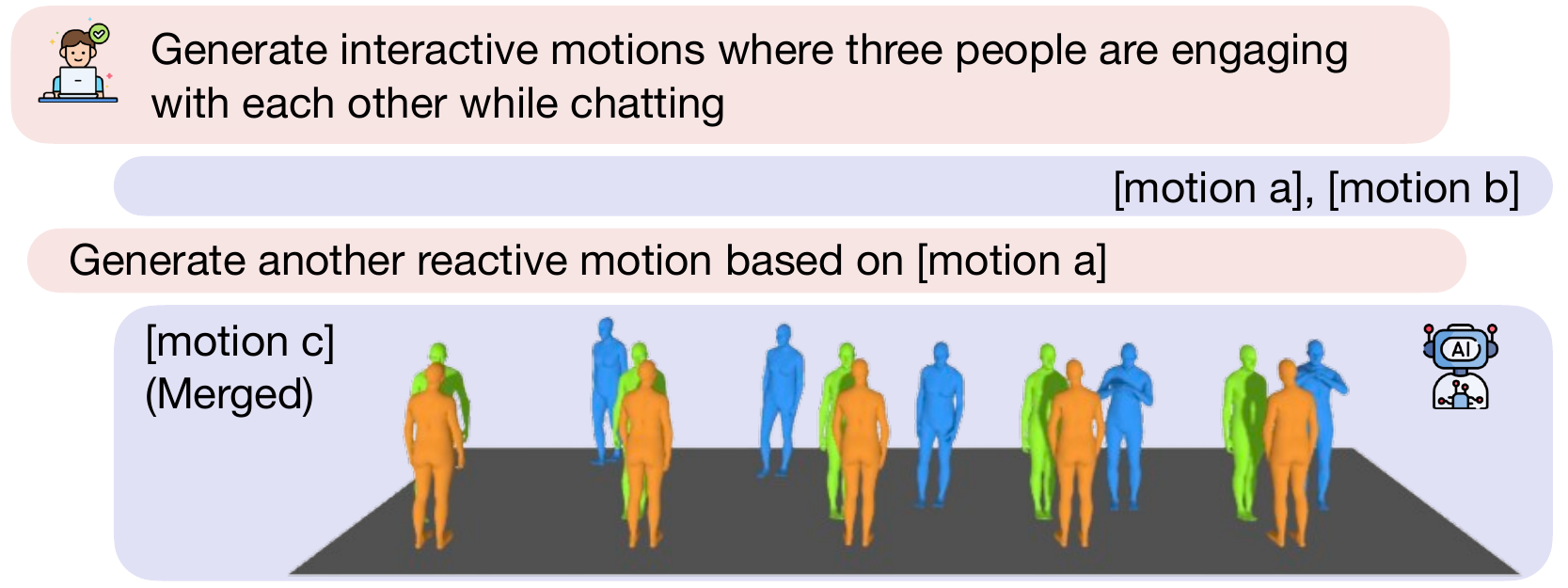}
    \caption{Our method demonstrates its extendability by generating multi-person interactions ($\geq 3$ people) through iterative prompting. Despite being trained on two-person scenarios, our framework conditions new motions on prior interactions, enabling the synthesis of natural group dynamics from textual descriptions.}
    \label{fig:three}
\end{figure}
Our method demonstrates the capability to extend motion generation to multi-person interactions ($\geq 3$ individuals) through iterative prompting, despite being trained exclusively on two-person motion scenarios. Specifically, we first generate motion for a pair of individuals based on the given text description, and then the motion of additional participants is synthesized while conditioning on the pre-existing interactions. This approach ensures that the newly generated motions remain coherent and contextually appropriate within the evolving group dynamics. We argue that this extendability is a key advantage of our versatile framework, as it allows for the scalable generation of complex human interactions without requiring additional multi-person training data. An example of this process is illustrated in Figure \ref{fig:three}, where the model successfully generates a realistic group conversation scene based on a textual prompt.

\section{Illustration on Data Collection Pipeline}\label{sec:5}
\begin{figure}[!h]
    \centering
    \includegraphics[width=0.8\linewidth]{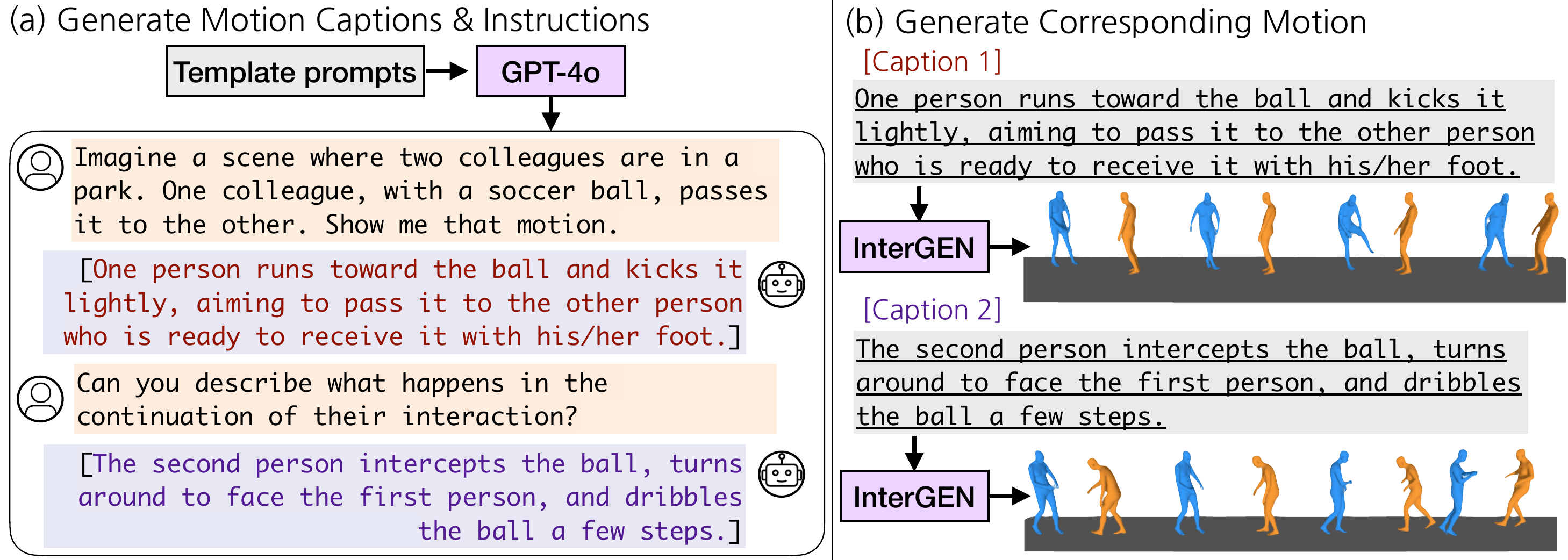}
    \caption{Overview of synthetic data generation for multi-turn conversations with interactive motions. (a) Motion captions and instructions are generated using GPT-4o based on interactions between two characters, followed by (b) the corresponding motion being synthesized using the InterGEN. }
    \label{fig:syn_pipeline}
\end{figure}
We illustrate the data collection pipeline for generating synthetic multi-turn conversations paired with interactive motions. As shown in Figure~\ref{fig:syn_pipeline}, GPT-4-based prompts are used to create captions and instructions, which are then converted into corresponding motions using InterGEN~\cite{liang2024intergen}.

\section{Additional Results for Traditional Motion Related Tasks}\label{sec:6}

\subsection{Motion to Text}
\begin{table}[!h]
    \centering
    \caption{Motion-to-Text}
    \begin{tabular}{c|ccc|c|c|c}
        \toprule
        Methods & \multicolumn{3}{c|}{Ret. Precision}  & \multirow{2}{*}{BLEU $\uparrow$} &\multirow{2}{*}{METEOR $\uparrow$} & \multirow{2}{*}{Rouge-L $\uparrow$} \\
        & Top1 $\uparrow$  & Top2 $\uparrow$  & Top3 $\uparrow$  & & &\\
        \midrule
        \textcolor{gray}{\textit{task-specific approach}} & & & & & \\
        TM2T$^*$  & 0.413  & 0.589 & 0.696 &  0.192 & 0.386 & 0.395\\
        \midrule
        \textcolor{gray}{\textit{unified approach}} & \\
        MotionGPT$^*$ &  0.288 & 0.405 & 0.494 & 0.000 & 0.000 & 0.00 \\
        MotionGPT$^*_I$ &  0.282 & 0.423 & 0.503 & 0.000 & 0.000 & 0.00 \\
        VIM-w/o \dataset & \textbf{0.677} & \underline{0.831} &  \underline{0.894} & \underline{0.220} & \underline{0.433} &  \underline{0.412}\\
        \textbf{VIM} (Ours) & \underline{0.669} & \textbf{0.842} &  \textbf{0.903} & \textbf{0.230} & \textbf{0.441} & \textbf{0.420} \\
        \bottomrule
    \end{tabular}%
    \label{tab:m2t}
\end{table}

Table \ref{tab:m2t} presents a comparative analysis of various methods on the motion-to-text generation task, focusing on retrieval precision and language evaluation metrics such as BLEU, METEOR, and Rouge-L. Among the task-specific approaches, TM2T$^*$ achieves moderate retrieval precision scores, with a Top1 precision of 0.413, and language metrics of 0.192 for BLEU and 0.386 for METEOR. In contrast, our proposed unified method, VIM, attains significantly higher retrieval precision scores, with a Top1 precision of 0.669, and surpasses TM2T* in all language metrics, achieving a BLEU score of 0.230 and a METEOR score of 0.441. This indicates that VIM not only narrows but effectively reverses the performance gap between task-specific and unified approaches in this task. The superior performance of VIM in both retrieval precision and language generation metrics demonstrates its effectiveness in generating accurate and descriptive textual captions from motion inputs. By achieving higher scores than the task-specific TM2T$^*$, VIM showcases the potential of unified approaches to not only close but surpass the performance gap traditionally observed between task-specific and unified models in motion-to-text generation tasks. This advancement underscores the ability of VIM to balance motion understanding and language generation, leading to more coherent and relevant textual outputs.


\subsection{Text to Motion}
\begin{table}[!h]
    \centering
    \caption{Text-to-Motion}    
    \begin{tabular}{c|ccc|c|c|c}
        \toprule
        Methods & \multicolumn{3}{c|}{Ret. Precision}
        & \multirow{2}{*}{FID $\downarrow$} & \multirow{2}{*}{Diversity $\rightarrow$} & \multirow{2}{*}{MMDist $\downarrow$}  \\
        & R Top1 $\uparrow$ & R Top2 $\uparrow$ & R Top3$\uparrow$  & & & \\
        \midrule
        Real & 0.649 & 0.807 & 0.878 & 0.00 & 0.988 &1.072 \\
        \midrule
        \textcolor{gray}{\textit{task-specific approach}} & & & & & \\
        TM2T$^*$ & 0.276 & 0.437 & 0.534 &  0.300 & 0.676 & 1.130 \\
        MoMask$^*$ & 0.402 & 0.535 & \underline{0.612}& \textbf{0.066} & \textbf{0.973} & \underline{ 1.128}\\
        ComMDM & 0.090 & 0.122 & 0.201 & 0.302 & 0.578 & 1.201 \\
        InterGEN & \textbf{0.403} & \textbf{0.557} & \textbf{0.645} & 0.078 & \underline{0.957} & \textbf{1.115}\\
        \midrule
        \textcolor{gray}{\textit{unified approach}} & & & & & \\
        MotionGPT$^*$ & 0.180 &  0.262 & 0.328 & 0.123 & 0.898 & 1.167 \\
        MotionGPT$^*_I$ &0.175 & 0.264 & 0.331& 0.118 & 0.900 & 1.176 \\
        VIM-w/o \dataset & 0.335 & 0.466 & 0.561 & 0.082 & 0.922 &1.127 \\
        \textbf{VIM}(Ours) & 0.318 &0.469 & 0.568 & \underline{0.059} & 0.945 & 1.126 \\
        \bottomrule
    \end{tabular}%
    \label{tab:t2m}
\end{table}
Table \ref{tab:t2m} provides a comparative analysis of various methods on the text-to-motion generation task, emphasizing the R Top3 retrieval precision metric. Among the task-specific approaches, MoMask* achieves the highest R Top3 score of 0.844, closely approaching the real data benchmark of 0.878, indicating its superior ability to retrieve relevant motions corresponding to textual inputs. InterGEN and TM2T* attain R Top3 scores of 0.645 and 0.534, respectively, showing moderate performance in capturing the top three relevant motions. In contrast, our proposed unified method, VIM, achieves an R Top3 score of 0.568, outperforming other unified methods like MotionGPT$^*$ and MotionGPT$^*_I$, which have lower R Top3 scores of 0.328 and 0.331, respectively. Although VIM does not surpass the task-specific MoMask in R Top3 precision, it narrows the performance gap between task-specific and unified approaches. Additionally, VIM maintains a favorable FID score of 0.059 and a high diversity of 0.945, suggesting that it effectively balances motion relevance with quality and variety. 
\subsection{Reaction Generation}
\begin{table}[!h]
    \centering
    \caption{Reaction Geneneration}  
    \begin{tabular}{c|c|c|ccc|c}
        \toprule
        \multirow{2}{*}{Methods} & \multirow{2}{*}{MPJPE $\downarrow$} & \multirow{2}{*}{FID $\downarrow$} & \multicolumn{3}{c|}{Ret. Precision} & \multirow{2}{*}{MMDist $\downarrow$} \\
        & & & R Top1 $\uparrow$ & R Top2 $\uparrow$ & R Top3 $\uparrow$  \\
        \midrule
        \textcolor{gray}{\textit{task-specific approach}} & \\
        MoMask$^*$ & 1.602 & 0.112 & 0.109 & 0.328 & 0.412 & 1.178\\
        \midrule
        \textcolor{gray}{\textit{unified approach}} & \\
        MotionGPT$^*$ & 3.441 & 0.355 & 0.079 & 0.104 & 0.355 & 1.246 \\
        MotionGPT$^*_I$ & 1.486 & 0.106 & 0.059 & 0.128 & 0.106 & 1.215\\
        VIM-w/o \dataset & \underline{0.984} & \underline{0.031} & \underline{0.311} & \underline{0.459} & \underline{0.554} & \underline{1.121} \\
        \textbf{VIM} (Ours) & \textbf{0.690} & \textbf{0.019} & \textbf{0.381} &  \textbf{0.537} & \textbf{0.625} & \textbf{1.110} \\
        \bottomrule
    \end{tabular}%
    \label{tab:regen}
\end{table}
Table \ref{tab:regen} presents a comparative analysis of various methods on the reaction generation task, focusing on metrics such as MPJPE, FID, Retrieval Precision, and MMDist. Our proposed unified approach, VIM, achieves the best performance across all evaluated metrics. Specifically, VIM attains the lowest MPJPE of 0.690 and the lowest FID of 0.019, indicating highly accurate joint position predictions and high fidelity in generated motions, respectively. In terms of retrieval precision, VIM outperforms both the task-specific method MoMask$^*$ and other unified approaches, achieving R Top1, R Top2, and R Top3 scores of 0.381, 0.537, and 0.625. Additionally, VIM has the lowest MMDist of 1.110, suggesting that it generates motions closest to the real data distribution. The ablation model, VIM-w/o \dataset, also performs well but slightly lags behind VIM, highlighting the significance of the \dataset component in enhancing performance. These results demonstrate that VIM not only narrows but effectively surpasses the performance gap between task-specific and unified approaches in reaction generation tasks, showcasing its effectiveness in generating accurate and realistic motion reactions.

\section{Limitations and Impact Statement}\label{sec:7}
The expressiveness of our models remains limited when handling complex or previously unseen actions, indicating a need for further diverse motion source data in its ability to generalize across diverse motion scenarios. 
In addition, the sequence length becomes excessively long as we flatten the residual motion tokens, which can impact efficiency and computational resources. Leveraging additional transformer models to predict the residual token can reduce this work. 
Lastly, our method faces challenges in personalization and interpretability, as motion is inherently ambiguous and users may interpret the same motion in different ways. Addressing this issue will require incorporating more tailored approaches that adapt to individual user preferences and expectations through further human-in-the-loop feedback and refinement processes. 
\paragraph{Broader Impact Statement}
Our method opens up new possibilities for interactive motion modeling and understanding, potentially benefiting fields like robotics, virtual environments, and human-computer interaction. However, as the model evolves, careful consideration of ethical concerns, such as misinterpretation of motions or unintended behavioral biases, is crucial. 

\section{Implementation Details}\label{sec:8}
\begin{figure}[!h]
    \centering
    \includegraphics[width=0.9\linewidth]{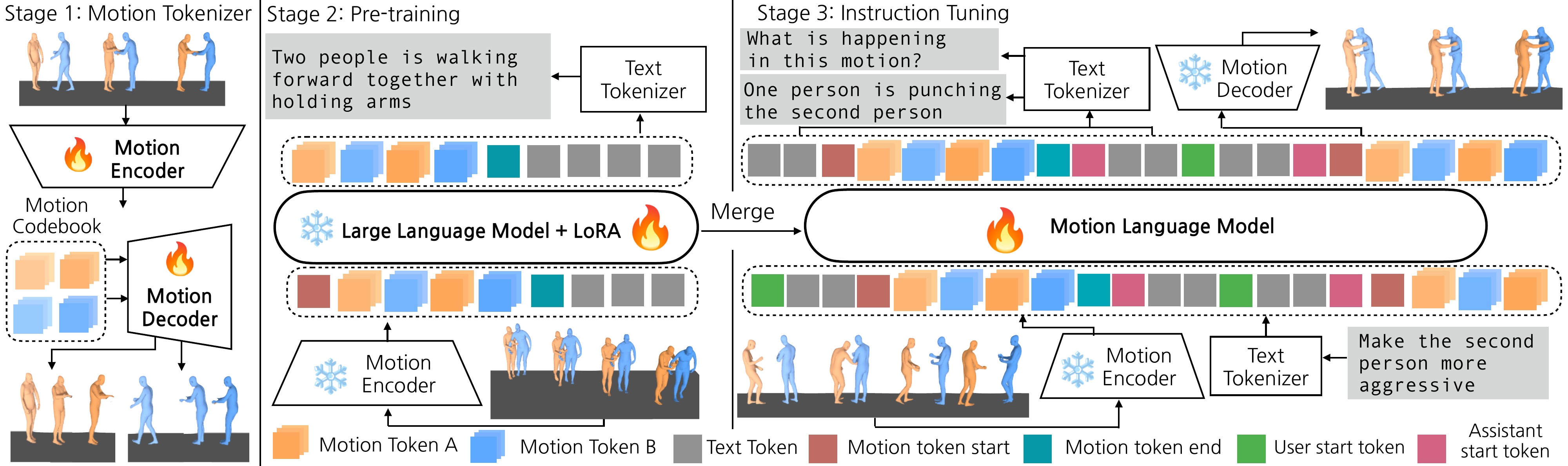}
    \caption{Method Overview. Stage 1 involves training a motion tokenizer that encodes and decodes interactive motion data. In Stage 2, we pre-train the model by integrating motion and text data, allowing it to learn the alignment between text and motion. Stage 3 focuses on Instruction Tuning, fine-tuning the model to follow instructions and improve its responsiveness to conversational cues.}
    \label{fig:method_appendix}
\end{figure}
We set the codebook of the motion tokenizer as $K \in R^{512 \times 512}$ for most
comparisons, with residual depth $4$. The motion encoder $\mathcal{E}$ incorporates a temporal downsampling rate $l$ of 4. We utilize LLaMA-3.1-8B~\cite{dubey2024llama} as the underlying architecture for our language model. During the pertaining, we train the large language model (LLM) using a low-rank adaptor (LoRA)~\citep{hu2022lora}, including the embedding layer and the decoder head. The rank was set as $r=8$, $\alpha=16$, with the dropout rate set as $0.05$. During the instruction fine-tuning stage, we trained all the parameters. The learning rate was set as $0.0001$, the warm-up ratio as $0.01$, the learning rate scheduler with cosine decay, and the AdamW optimizer. For single-person modeling, we feed the encoder two identical copies of the motion. In other words, we concatenate the motion with itself so that the encoder, which expects two inputs, processes the same motion twice.

In this implementation, we first construct a motion tokenizer by training a motion encoder–decoder pair (Stage 1), where the encoder converts raw motion data into discrete tokens and the decoder can accurately reconstruct these motions from the tokens. Next, during pretraining (Stage 2), we fuse text tokens (from a large language model) with the motion tokens to jointly model their relationship, ensuring that the textual representations align with the motion embeddings. Finally, in the instruction-tuning phase (Stage 3), we refine the entire model using interactive text–motion tasks. Here, the model is trained to follow higher-level instructions by leveraging both the textual context and the learned motion tokens, enabling it to generate context-appropriate motions in response to user queries or commands.
Figure~\ref{fig:method_appendix} illustrates the overall pipeline of the proposed method. 

\section{Implementation details for Baselines$^*$}\label{sec:9}
For training MotionGPT~\cite{jiang2023motiongpt}$^*$, TM2T~\cite{guo2022tm2t}$^*$ and MoMASK~\cite{guo2024momask}$^*$ in the interactive motion dataset, we have utilized the Flan-T5-base model~\cite{chung2024scaling} as a base large language model. We trained the model with Interhuman~\cite{liang2024intergen} and InterX~\cite{xu2024inter} dataset, with the non-canonical representation, the same as the proposed method. 
To model the interactive motion, motion tokens of a person ``A" and ``B" are fed interleaved such as,
\begin{equation}
    \texttt{<motion\_token\_start>}, k^{1;a}, k^{1;b}, k^{2;a}
\end{equation}, where $k^{i;a}$ is the $i$-th token of motion $a$. 
Although scaling up the model can improve the performance, we conducted the experiment with the same base model as the original paper from MotionGPT~\cite{jiang2023motiongpt} and Motionchain~\cite{jiang2024motionchain}. The original paper reported that increasing the model size did not significantly improve the model's performance. 
We followed same motion token representation as MotionGPT$^*$ for both TM2T~\cite{guo2022tm2t}$^*$ and MoMASK~\cite{guo2024momask}$^*$.

\section{Detailed Task Explanations}\label{sec:10}

\paragraph{Motion Editing} 
Standard motion editing tasks typically involve modifying the motion of a single person based on physical descriptions, such as "raise higher" or "move faster." However, in this task, we focus on editing interactive motions involving two people based on their personas, such as emotions or relationships, by modifying just one person's persona. The primary challenge in motion editing for two people is that when the motion of one person changes, the motion of the second person, which is correlated, also needs to be adjusted. This requires more complex reasoning about social interactions. Specifically, we define the task as ``\texttt{USER:\{scene\_information\}, \{reference\_motion\}. ASSISTANT: \{motion\_caption\}. USER: \{editing\_command\}. ASSISTANT: \{edited\_motion\}.}" The editing command could be defined as asking the model to change the persona of a person, like ``Make one person shy." We let our model generate motion caption in the middle to let the chain-of-thoughts technique improve the reasoning ability. 

\paragraph{Motion Reasoning}
Motion reasoning involves predicting future motions or inferring past events based on the current motion context. This task requires understanding the sequence of motions and making logical inferences about the preceding or subsequent events. For instance, given a motion of an ongoing interaction between two individuals, the model needs to deduce what might have happened before this moment or predict what will likely occur next. 
This is crucial for applications requiring a temporal understanding of motions, such as surveillance analysis, animation, or human-robot interactions. We define the input sequence as follows: ``\texttt{USER:\{question\_1\}, \{motion\_1\}. ASSISTANT: \{answer\_1\}. USER: \{question\_2\}, \{motion\_2\}.}", where the model has to predict ``\texttt{ASSISTANT: \{answer\_2\}}".
The inference question could involve queries like "Can you tell me what happened before?" or "What do you think will happen next in this scenario?". This task demands high-level reasoning and comprehension of motion sequences, enabling the model to generate plausible and coherent motion narratives based on the given context.

\section{Detailed Explanation about Two-stage Baselines}\label{sec:11}

In Section 5.2 and Section 5.3, we have compared the proposed method with two-stage models. In particular, we have utilized TM2T~\cite{guo2022tm2t} for the motion captioner and InterGEN~\cite{liang2024intergen} for the text-to-motion generation model.

\subsection{Motion Editing}
In the motion editing task, the two-stage approach first uses the motion-to-text (TM2T; \cite{guo2022tm2t}) model to generate a caption from the source motion and append the editing command. Then, the text-to-motion (InterGen; \cite{liang2024intergen}) model produces the edited motion based on this caption and command. In particular, the input for the text-to-motion model is ''\texttt{[motion caption]. [editing command]}".

We first trained the TM2T model with the InterHuman dataset~\cite{liang2024intergen}and the InterX~\cite{xu2024inter} dataset, similarly to MotionGPT$^*$.   
Next, we trained the text-to-motion diffusion model, InterGEN for the second stage. 

\subsection{Motion Reasoning}
In the motion reasoning task, the two-stage model integrates TM2T with large language models such as GPT-4o~\cite{openai244o} and LLaMA-3.1-8B~\cite{dubey2024llama}. Here, the motion components in the conversational data are replaced with captions generated by TM2T, which are then fed into the LLM for reasoning and response generation. In particular, the original input for the motion-language model was ``\texttt{USER:\{question\_1\}, \{motion\_1\}. ASSISTANT: \{answer\_1\}. USER: \{question\_2\}, \{motion\_2\}.}", where the model has to predict ``\texttt{ASSISTANT: \{answer\_2\}}". We replaced the motion into motion caption obtained by motion captioner for the input for LLM like ``\texttt{USER:\{question\_1\}, \{motion\_caption\_1\}. ASSISTANT: \{answer\_1\}. USER: \{question\_2\}, \{motion\_caption\_2\}.}". Again, we utilized TM2T$^*$ for the motion captioner mentioned in the previous section.

\section{More details about Evaluation Metric for Traditional Motion Related Tasks}\label{sec:12}

\paragraph{Motion Quality}
The Frechet Inception Distance (FID) is used to assess the similarity between the distributions of generated and real motions, utilizing an appropriate feature extractor tailored to each dataset. In addition, we use well-known motion capture metrics, MPJPE to quantify global and local errors in meters.
\paragraph{Motion Diversity}
We have utilized diversity to evaluate the diversity of the motion following previous work \citep{jiang2023motiongpt,petrovich23tmr}. 
To evaluate Diversity, the generated motions are split into two equal-sized subsets, and the Diversity metric is calculated as the average distance between motions within these subsets.
\paragraph{Text-Motion Matching}
TMR \citep{petrovich23tmr} offers motion/text feature extractors that produce geometrically coherent features for aligned text-motion pairs and vice versa. In this feature space, we evaluate motion-retrieval precision (R Precision) by combining the generated motion with 31 mismatched motions and calculating the top-1/2/3 matching accuracy between the text and motion. Furthermore, we assess the Multi-modal Distance (MM Dist), which measures the distance between the generated motions and their corresponding text.

\section{Template Forms for Pre-training and Instruction Tuning}\label{sec:13}

\begin{table*}[]
    \centering
    \caption{Template for Pretraining}
    \begin{tabular}{ccc}
        \toprule
         Task& Sequence & Label \\
         \midrule
         Text-to-Motion & Generate caption from motion: [motion] [caption] & [caption] \\
         Motion-to-Text & Generate motion from caption: [caption][motion]  & [motion]\\ 
         Reaction Generation & Generate reaction motion: [motion] & [motion B] \\
         Motion Prediction & Predict motion: [motion] & [Last 75\%motion]\\
         \bottomrule
    \end{tabular}
    \label{tab:pre}
\end{table*}

\begin{table*}[]
    \centering
    \caption{Template for Instruction Tunning}
    \resizebox{1\textwidth}{!}{
    \begin{tabular}{ccc}
        \toprule
         Task& User & Assistant \\
         \midrule
         Text-to-Motion & Demonstrate a sequence of movements that symbolizes the sentiment of [caption] & [motion]\\ 
          & Please create a motion that represents the power of  [caption] & The motion is [motion]\\ 
          & I need a motion that represents the power of [caption] & Sure, [motion]\\
          &Show me a gesture that conveys [caption] & \\
          & Produce a motion that matches [caption] &\\ 
           & $\cdots$ \\
          \midrule
           Motion-to-Text & Describe the motion represented by [motion] & [caption] \\
           & Provide a summary of the action depicted in [motion] & \\
           &Explain the motion shown in [motion] & \\
           & Provide a text-based explanation of the action being shown in [motion] & \\
           & Please provide a description of the motion in [motion] & \\
           & $\cdots$ \\
           \midrule
         Motion Prediction & Predict motion: [first 25\%motion] & [Last 75\%motion]\\
        & Do the motion prediction task for [first 25\%motion] & \\
         \bottomrule
    \end{tabular}}
    \label{tab:inst}
\end{table*}
We will detail the template forms utilized during the pre-training and instruction-tuning stages of our model development. Tables \ref{tab:pre} and \ref{tab:inst} illustrate the specific formats employed in each stage, providing a structured approach to aligning motion data with textual descriptions and enhancing the model's interactive capabilities. All the templates are from MotionGPT~\citep{jiang2023motiongpt}.

\subsection{Pre-training Templates}

During the pre-training stage, our objective is to align motion and language representations by leveraging large language models (LLMs). We design tasks such as Text-to-Motion, Motion-to-Text, Reaction Generation, and Motion Prediction using paired datasets like InterX~\cite{xu2024inter} and Interhuman~\cite{liang2024intergen}. The pre-training templates involve generating captions from motion sequences, creating motions based on textual descriptions, producing reaction motions in response to initial motions, and predicting subsequent motions from partial sequences, as summarized in Table \ref{tab:pre}. For single-person motion, we utilized text-to-motion, motion-to-text and motion prediction task during training. 

\subsection{Instruction-Tuning Templates}

In the instruction-tuning stage, we enhance the model's ability to follow diverse instructions presented in a conversational format. Utilizing the INTER2-MT dataset alongside single-turn data from previous interactive motion datasets, we format user instructions and assistant responses to facilitate multi-turn interactions. Table \ref{tab:inst} outlines the templates used for tasks such as generating motions from user prompts, describing motions based on user queries, and predicting motion continuations. By structuring the interactions in this manner, the model becomes adept at understanding and responding to various motion-related commands, thereby improving its performance in interactive scenarios.


\section{Data Sample Visualization}\label{sec:14}
\begin{figure*}[!h]
    \centering
\includegraphics[width=0.8\linewidth]{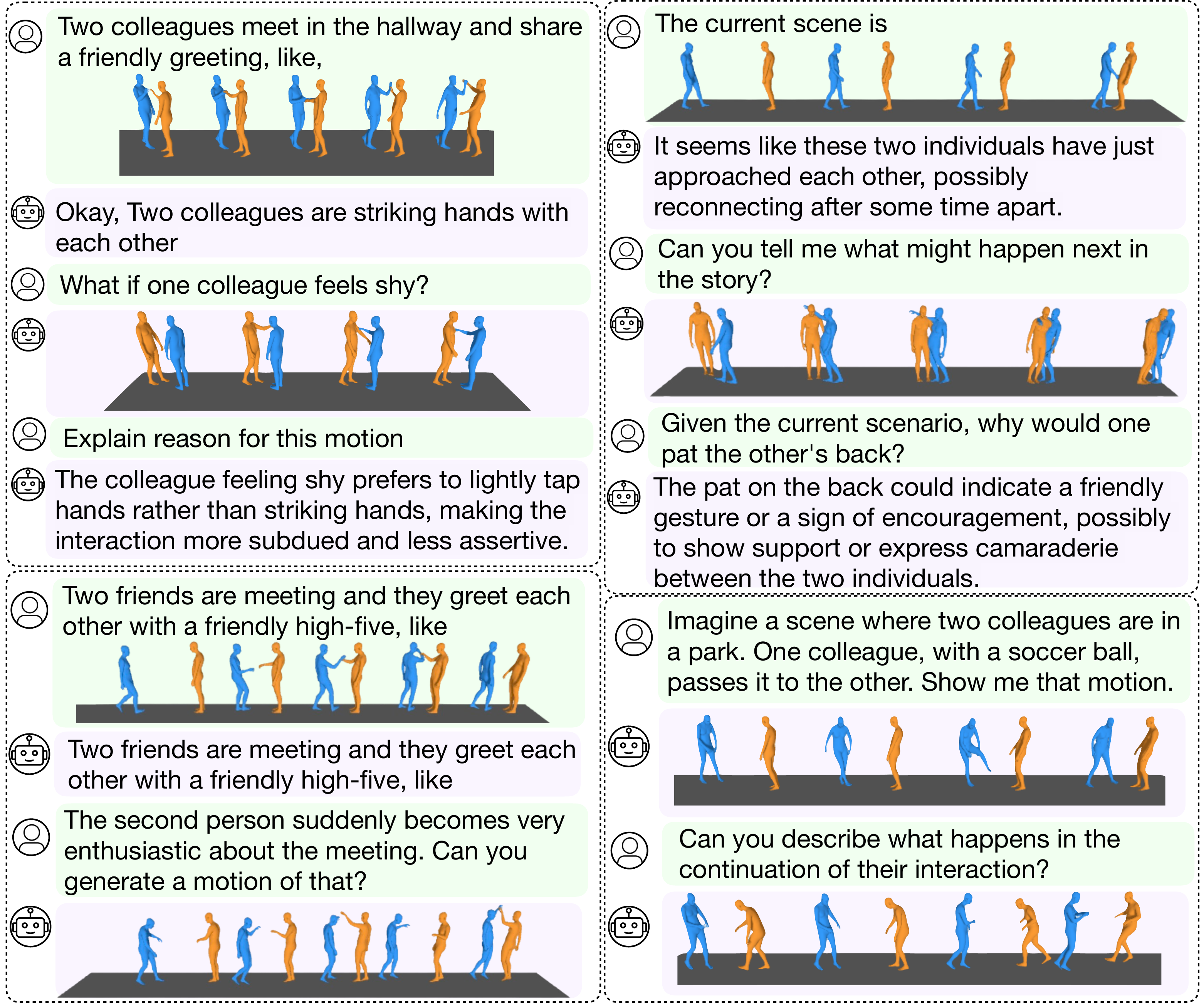}
    \caption{Sample from \dataset~dataset. The left column visualizes samples of motion editing, and the right column shows examples from the motion reasoning task. }
    \label{fig:data}
\end{figure*}

The samples from the synthesized dataset, \dataset, are illustrated in Figure \ref{fig:data}. 

\section{\dataset~Statistics}\label{sec:15}

\begin{table}[!h]
    \centering
    \caption{Statistics on \dataset. }
   \begin{tabular}{ccccc}
        \toprule
             &  Total & Train & Val. & Test\\
            \midrule
            \# of Samples & 82736 & 66194 & 4141 & 12401 \\
            \midrule
            \# of Motions &  317749 &132388 & 8282 & 24802  \\
             From Dataset & 56395 &50258 & 3142   & 2995 \\
             Synthesized & 96676 &82130 & 5140 & 9406 \\
            \bottomrule
        \end{tabular}
    \label{tab:my_label}
\end{table}
%
We collected 82K samples of multi-turn conversational data, each involving interactive motions. Of these, 30K samples focus on motion editing, 30K on reasoning about past or future scenarios, and 12K on story generation. Each sample includes four to eight conversation turns and two distinct motions. The dataset contains 96K motions generated using a text-to-motion diffusion model, while 56K motions come from the original source dataset. The train-validation-test set is randomly split by the ratio 0.8:0.05:0.15. 


\clearpage
\section{Qualitative Results}\label{sec:16}

\begin{figure*}[!h]
    \centering
\includegraphics[width=0.8\linewidth]{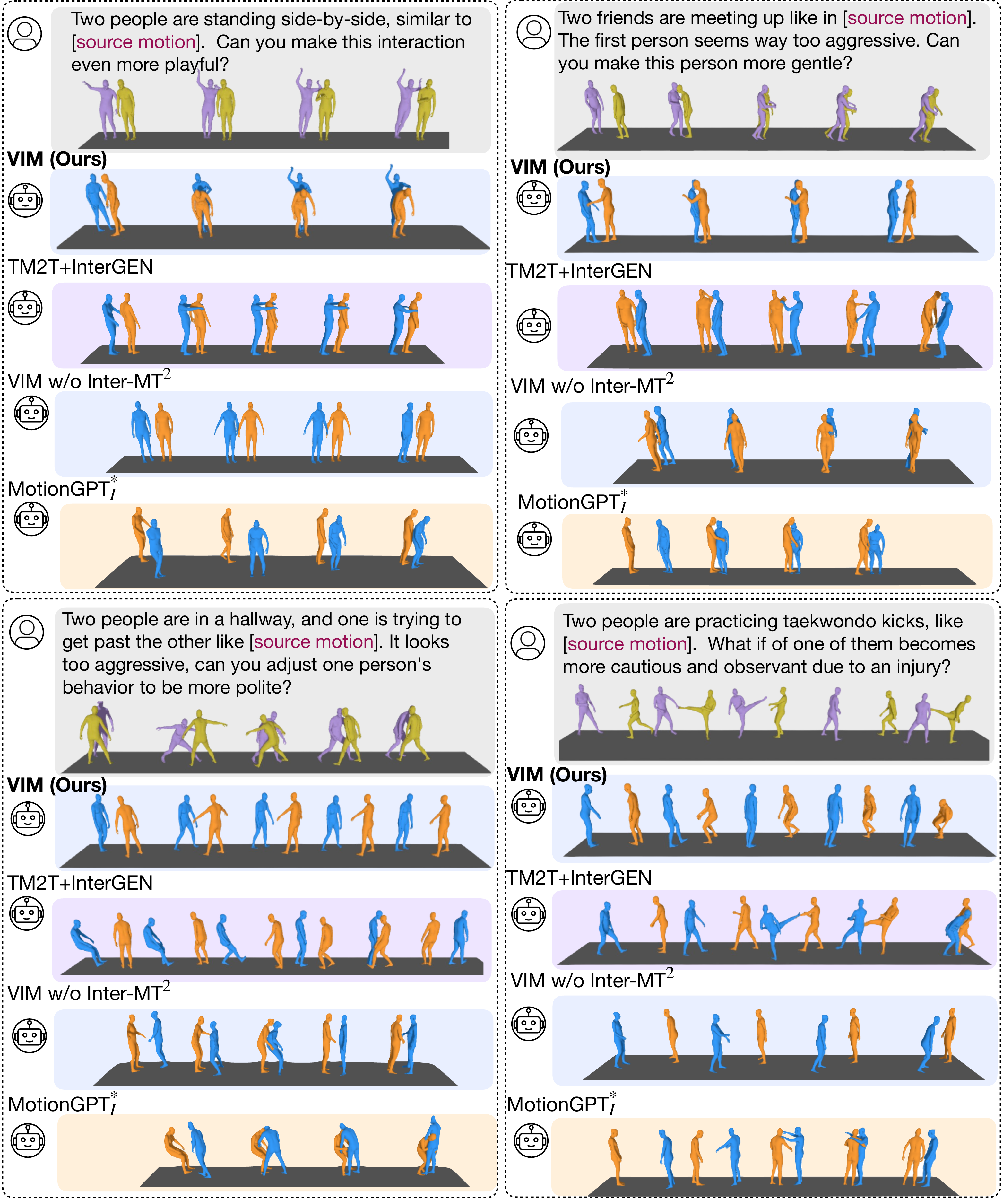}
    \caption{Generated motion from source and editing command. }
    \label{fig:demo_edit}
\end{figure*}

\begin{figure*}[!h]
    \centering
\includegraphics[width=0.8\linewidth]{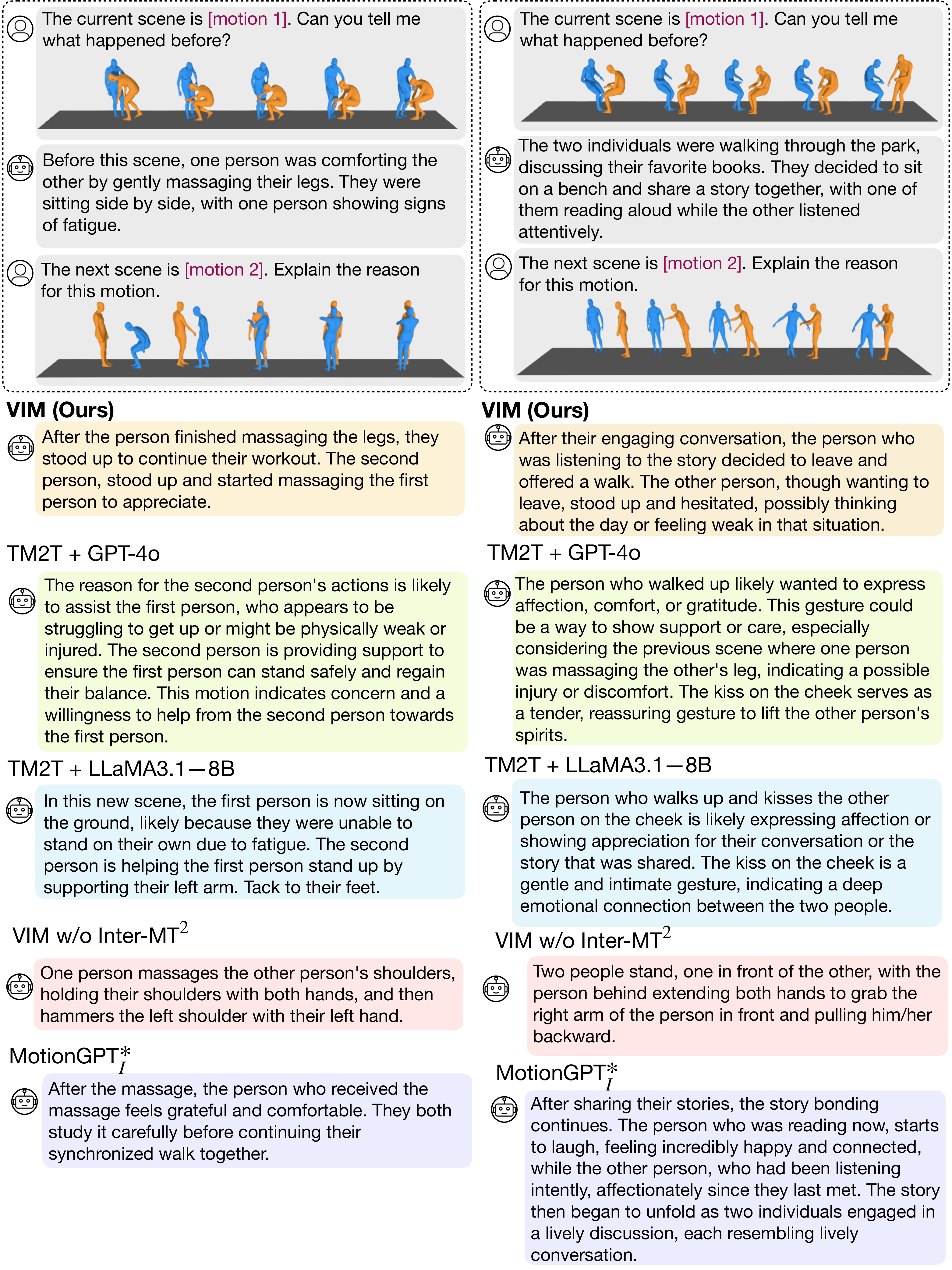}
    \caption{Generated responses based on the previous conversations for motion reasoning task. }
    \label{fig:demo_rea}
\end{figure*}

\begin{figure*}[!h]
    \centering
    \includegraphics[width=0.8\linewidth]{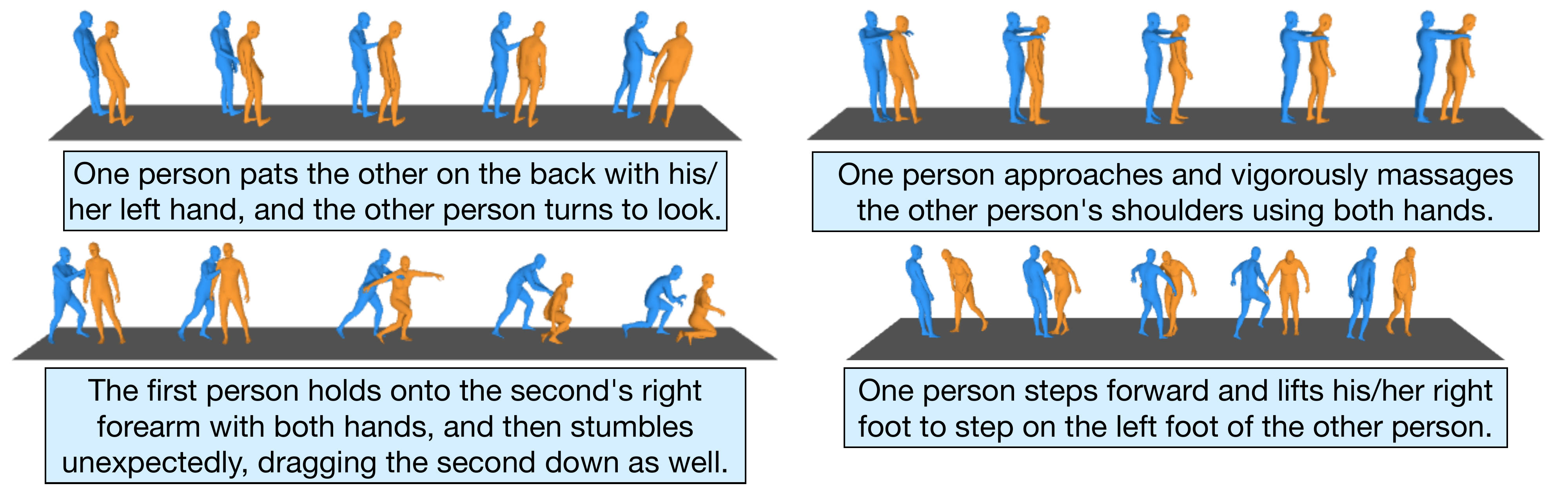}
    \caption{Motion-to-text results. The blue part is generated motion captions from source motions. }
    \label{fig:m2t}
\end{figure*}

\begin{figure*}[!h]
    \centering
    \includegraphics[width=0.8\linewidth]{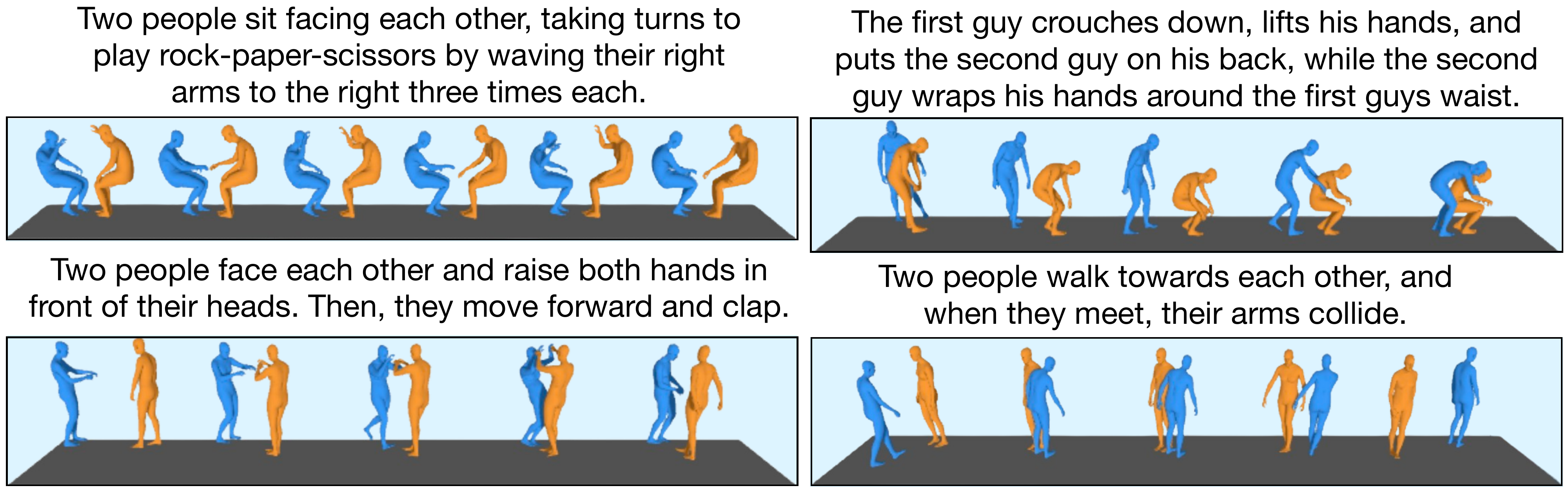}
    \caption{Text-to-motion results. The blue part is generated motions from the motion caption. }
    \label{fig:t2m}
\end{figure*}

\begin{figure*}[!h]
    \centering
    \includegraphics[width=0.8\linewidth]{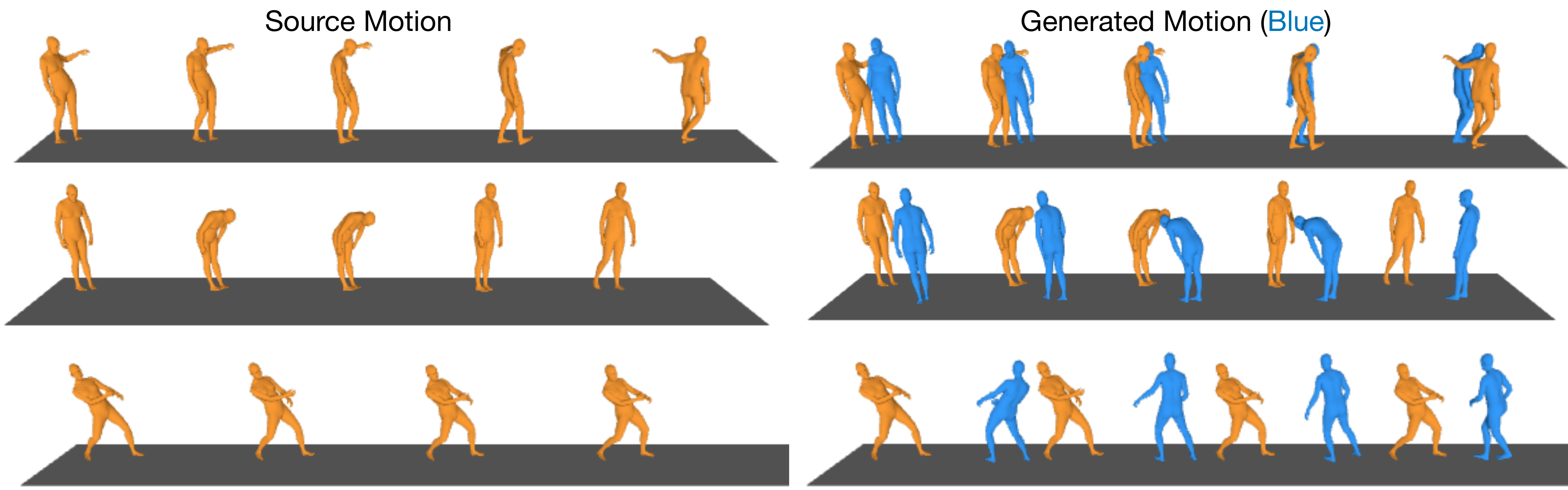}
    \caption{Reaction Generation. The input motion is orange, while the generated reactive motion is colored blue. }
    \label{fig:react}
\end{figure*}

\begin{figure*}[!h]
    \centering
    \includegraphics[width=0.6\linewidth]{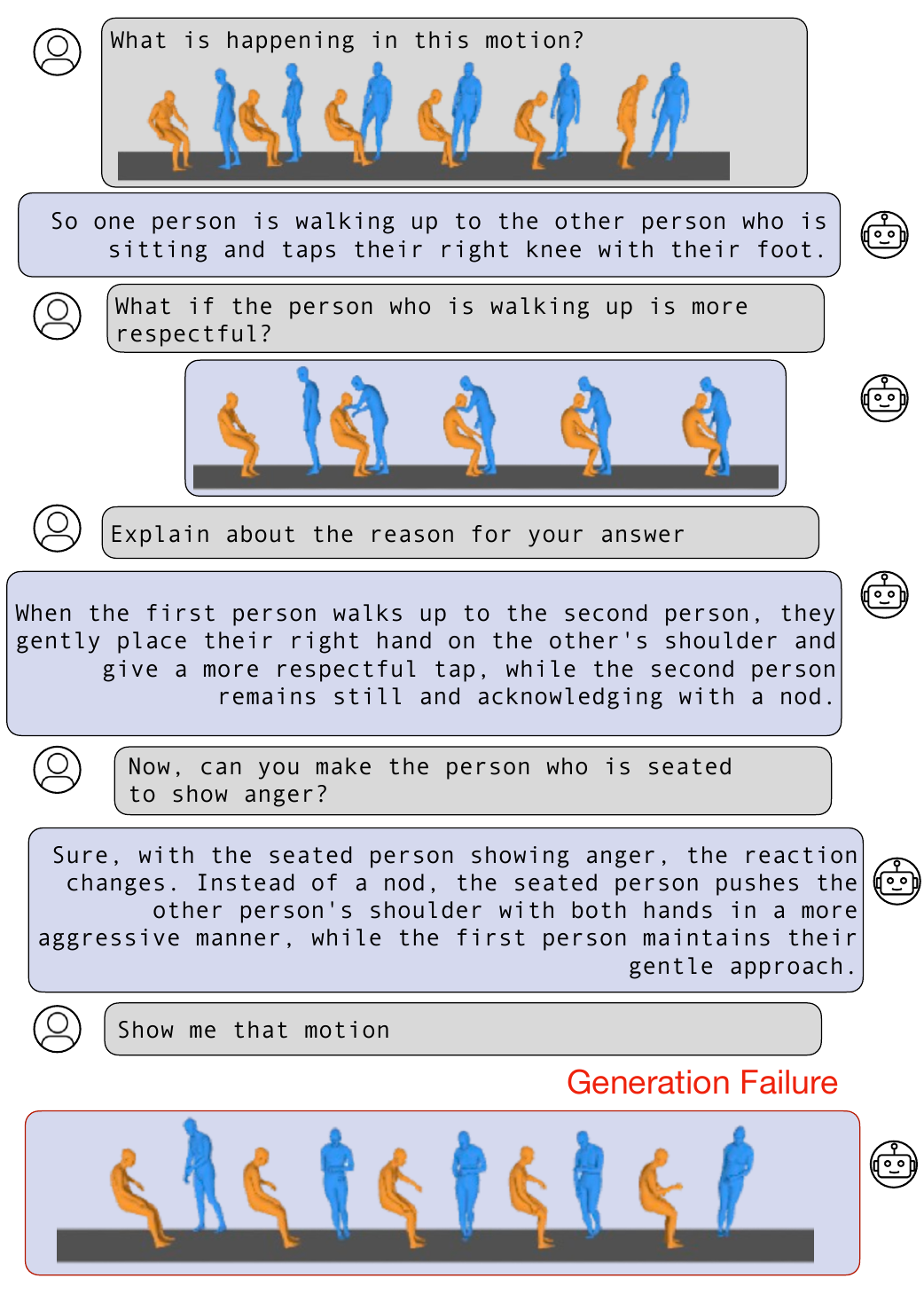}
    \caption{We demonstrate the ability of the proposed method to generate responses in long conversations and also the failure cases. }
    \label{fig:long}
\end{figure*}

We visualize our result gallery on motion editing in Figure \ref{fig:demo_edit} and on motion reasoning in Figure \ref{fig:demo_rea}. 
Furthermore, the results for motion-to-text (Figure \ref{fig:m2t}), text-to-motion (Figure \ref{fig:t2m}), and reaction generation (Figure \ref{fig:react}) are demonstrated. In figure \ref{fig:long}, we demonstrated the generation ability of the proposed method in longer contexts with a failure case.



\clearpage

\section{User Subject Studies Protocols for motion editing}\label{sec:17}
We conducted user subject studies using the platform on the Mechanical Turk service from AWS~\citep{mtruk}. 

\subsection{Instructions}
The summary given to the user is as follows:
\greybox{
We are conducting an academic survey about the quality of generated motions. 
We need to understand your opinion about the motion quality and ability to follow the editing commands. 
Please evaluate each motions based on the given criteria.

You will be presented with multiple instruction samples. 
After completing the evaluations on each page, click "Next" to proceed.
On the last page, click "Submit" to complete the survey.
}

The detailed instruction is as follows:
\greybox{
Objective:
We are conducting a survey to evaluate how well AI-generated motions follow given instructions and how natural they appear. Your feedback is important to help us improve the AI's ability to create realistic movements that match specific editing commands.

Survey Overview:
You will be shown a source motion and an edited motion. Your task is to evaluate both based on specific criteria. After evaluating a few examples, you will also rate multiple edited motions generated from the same source motion using different methods. The survey is divided into multiple pages, and you can move through the pages using "Next" or "Previous" buttons. You must complete all fields on each page before proceeding.

Evaluation Criteria:
For each pair of videos (source and edited), you will be asked to rate them based on:

Content Similarity:
Does the edited motion stay true to the original motion?
Rating scale: 1 (Strongly Disagree) to 5 (Strongly Agree)

Alignment with Instruction:
Does the edited motion follow the instructions given?
Rating scale: 1 (Strongly Disagree) to 5 (Strongly Agree)

Motion Quality:
Is the quality of the edited motion good, and does it look natural?
Rating scale: 1 (Strongly Disagree) to 5 (Strongly Agree)

Survey Structure:

Evaluation of Pre-selected Motion Examples:
In the first section, you will review hand-picked video pairs. Each page will show a source video and its edited version. You will rate how similar they are, how well the editing follows instructions and the overall quality of the motion.

Evaluation of Randomly Selected Motion Samples:
In the second section, you will see five different edited motions for each scenario. These motions are created using different methods. You will rate each one based on content similarity, alignment with instructions, and motion quality.

Instructions:

Review the motion examples:
Each page will show a description, editing instruction, and two videos (source and edited). Watch the videos and rate them using radio buttons based on the three criteria. Click "Next" to move to the next example.

Evaluate random scenarios:
You will be shown five edited motions per scenario. Review and rate them on the same criteria as before. Use "Next" and "Previous" to navigate.

Completion:
Once all evaluations are finished, click "Submit" to complete the survey.

Tips:

Watch both videos completely before deciding.
If you’re unsure, select "Neutral."
All fields must be filled before you can move forward or submit the survey.
}

\begin{figure}[!h]
    \centering
    \includegraphics[width=\linewidth]{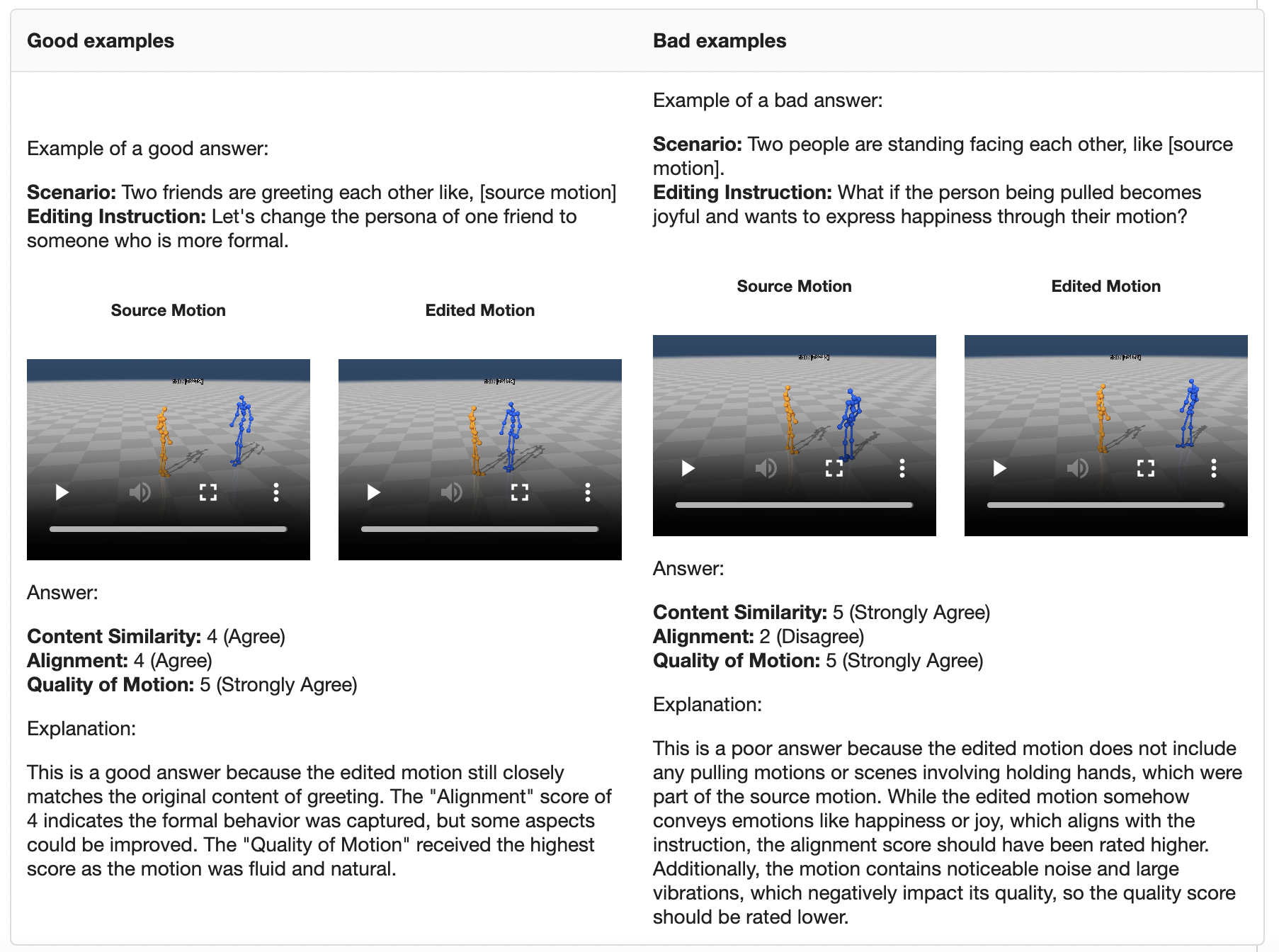}
    \caption{The examples of ratings given to the user}
    \label{fig:user_ex}
\end{figure}

The examples of ratings given to the user are shown in Figure \ref{fig:user_ex}. 

\subsection{Qualifying test}
\begin{figure}[!h]
    \centering
    \includegraphics[width=\linewidth]{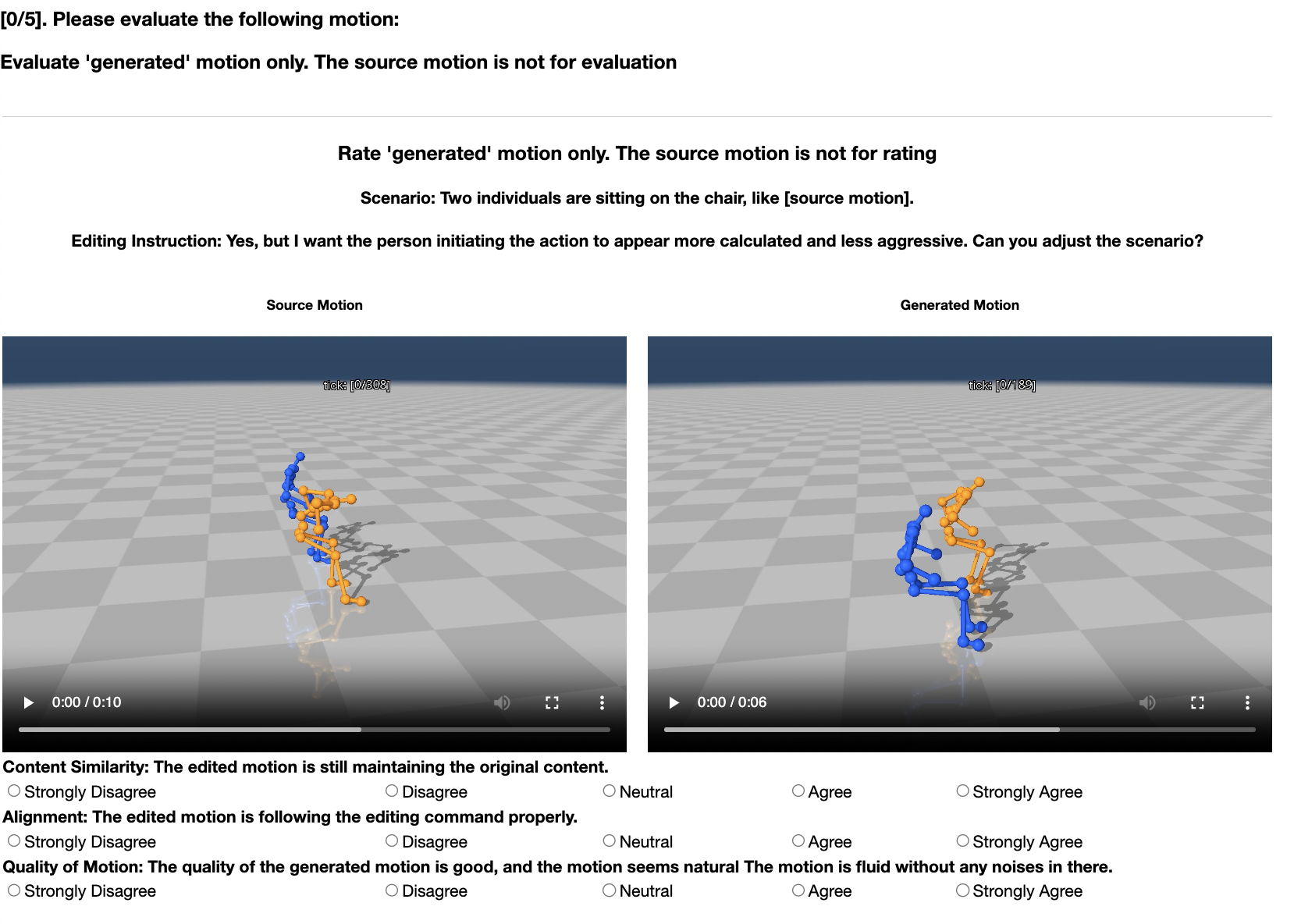}
    \caption{Qualifying test in user subject studies}
    \label{fig:user_qual}
\end{figure}

Before participating in the main user studies, all participants must pass a qualifying test to ensure they understand the evaluation criteria. In this test, participants are asked to assess four samples based on three metrics: Content Alignment, Fidelity of Motion, and Quality of Motion. Among the four samples, two are high-quality and derived from the ground-truth dataset, while the other two are low-quality—one is a mismatched motion with a single instruction, and the other is generated by the least effective model, MotionGPT$^*$. Participants must rate the low-quality samples lower than the high-quality ones in each of the three metrics. If any of the low-quality samples receive ratings that are equal to or higher than the high-quality samples in Content Alignment, Fidelity, or Quality of Motion, the participant will receive an error message and will need to adjust their ratings accordingly. This ensures that only participants who can accurately distinguish between high and low-quality motions based on the defined metrics proceed to the main study. The example of the qualifying test is demonstrated in Figure \ref{fig:user_qual}

\subsection{Detailed Survey Format}

\begin{figure}[!h]
    \centering
    \includegraphics[width=\linewidth]{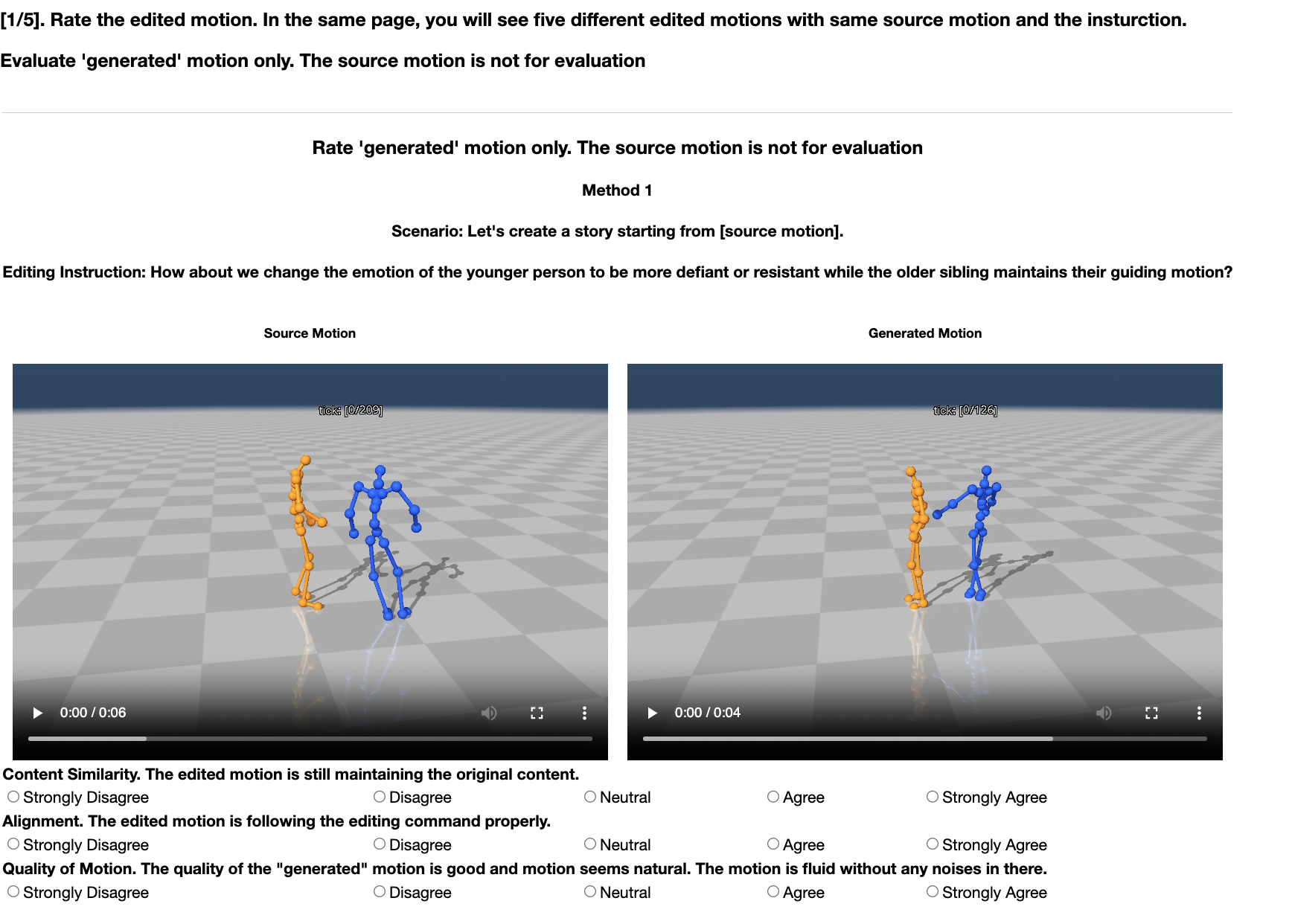}
    \caption{Survey Example}
    \label{fig:user_studies}
\end{figure}
\paragraph{Main Survey Structure}
In the main survey, each participant was randomly assigned 5 samples from a larger pool of 30 diverse motion sequences. This random sampling strategy was employed to ensure a broad and representative evaluation, minimizing any potential selection bias. For each of these selected samples, participants were asked to evaluate five baseline methods, including our proposed model (VIM), VIM w/o \dataset, MotionGPT$^*$, MotionGPT$^*_I$, and two-stage model based on TM2T~\citep{guo2022tm2t} and InterGEN~\citep{liang2024intergen}.
To eliminate ordering effects and ensure that the evaluation was solely based on the quality of the motions rather than their presentation order, the order of the baseline methods was randomly shuffled for each participant. This randomization was crucial in preventing any unintended bias that might arise from the sequence in which the methods were presented.
\paragraph{Evaluation Metrics} Participants assessed each motion sample using three evaluation metrics, which provided a multidimensional view of each model's performance:
\begin{itemize}
    \item Content Similarity: The edited motion is still maintaining the original content.
    \item Alignment with Instruction: The edited motion is following the editing command properly.
    \item Motion Quality: The quality of the generated motion is good, and the motion seems natural The motion is fluid without any noises in there.
\end{itemize}
We leveraged a 5-scale Likert scale, 1 from strongly disagree to 5 for strongly agree. 

\paragraph{Exclusion Criteria}

To maintain high data quality and ensure meaningful results, we implemented strict exclusion criteria. Participants who assigned the same rating across all evaluation metrics for every sample were excluded, as such uniformity indicated a lack of genuine engagement or understanding of the evaluation process. Additionally, those who provided identical ratings across all comparison methods for a given sample were also omitted. This approach ensured that only participants who thoughtfully differentiated between the methods based on their performance were included in the final analysis. These exclusion rules were essential in filtering out unreliable data and ensuring that the survey results accurately reflected the participants' true assessments of each model's performance.

\clearpage
\section{Prompts for Data collection in \dataset}\label{sec:18}
We have utilized two different prompts in the data collection pipeline. One is generating two different motion captions with conversational data. The other one is generating one motion and conversational data based on the sample motion and corresponding caption from the base dataset, Inter-X~\citep{xu2024inter} and InterHuman~\cite{liang2024intergen}. 

Motion editing prompts without base samples is constructed as follows:

\greybox{
You are an AI visual assistant, and you are seeing a motion.
Design a conversation between you and a person building a conversation about editing this motion.
In conversations, you should indicate who said using "User:", and"AI:" in the beginning but these two words
do not occur in sentences.
The answers should be in a tone that an AI visual assistant is seeing the motion and answering the
question. The scenario should always contain two people in the scene.
Generate a conversation about building a story from two different motions. 
The flow of the conversation is as follows:
1. Creating a scenario. REMBER to make a story in this.
2. Change the emotion or persona of just one person.
3. Describe how the motion will be changed, with one person maintaining the same motion.
"""Example:
    User: Let's create a story starting from [Two individuals sitting across from each other, with one person extending his/her left hand and the other person extending their left hand. They proceed to participate in a wrist-wrestling competition]. 
    AI: Two people are doing an arm-wrestling match, and each person is grabbing the right hand of the other person while sitting. 
    User: The next scene is [Two individuals sit across from each other, with one person extending his/her left hand and the other person extending both hands. They proceed to participate in a wrist-wrestling competition, where the second person utilizes both hands in an attempt to defeat the first person's left hand.].
    AI: The one person kept losing the game, which made him competitive to win the game.""",
"""Example:
    User: Two friends are doing an arm-wrestling match.
    AI: [Two individuals sit across from each other, with one person extending his/her left hand and the other person extending left hand. They proceed to participate in a wrist-wrestling competition]
    User: One person got competitive.
    AI: [Two individuals sit across from each other, with one person extending his/her left hand and the other person extending both hands. They proceed to participate in a wrist-wrestling competition, where the second person utilizes both hands in an attempt to defeat the first person's left hand.].
    User: Explain the reason for the motion.
    AI: The one person kept losing the game, which made him cheat to win the game.""",
"""Example:
    User: Two friends are doing an arm-wrestling match, like [Two individuals sit across from each other, with one person extending his/her left hand and the other person extending left hand. They proceed to participate in a wrist-wrestling competition]. 
    AI: Two people are doing an arm-wrestling match, each person is grabbing the right hand of the other person, while sitting. 
    User: The one person kept losing the game, which made him competitive to win the game. Can you generate a motion of what would happen then?
    AI:  [Two individuals sit across from each other, with one person extending his/her left hand and the other person extending both hands. They proceed to participate in a wrist-wrestling competition, where the second person utilizes both hands in an attempt to defeat the first person's left hand.]""",
"""Example:
    User: Let's start making a story. Two friends are doing an arm-wrestling match, like [Two individuals sit across from each other, with one person extending his/her left hand and the other person extending their left hand. They proceed to participate in a wrist-wrestling competition]. 
    AI: The one person kept losing the game, which made him competitive to win the game.
    User: Sounds interesting. Can you visualize it?
    AI: [Two individuals sit across from each other, with one person extending his/her left hand and the other person extending both hands. They proceed to participate in a wrist-wrestling competition, where the second person utilizes both hands in an attempt to defeat the first person's left hand.]"""
===========
Example format for the [motion caption]:
    - One person approaches, raises his/her right hand to grab the other person's right forearm, places his/her left hand on it, and walks in the direction the grabbed person is facing.
    - Two people face each other, one person lifts his/her right leg and walks towards the other person, stopping half a meter away.
    - A person falls and braces himself/herself on the ground with his/her right hand. Another person approaches, squats down, and grabs his/her left arm with both hands to assist him/her in standing up.
The content inside the bracket ([]) is a caption for the motion. This is for visualizing the motion, which is not given in textual form during inference. 
I will denote this as [motion caption]. 
Please denote [motion caption] when AI or the user has to answer in the motion sequence.

Please make [motion caption] that is similar to the following action labels: \vio{[Action LABELS]}, and other motions like everyday routines (e.g., passing objects, greeting, communicating, etc.), and professional motions (e.g., Taekwondo, Latin dance, boxing, etc.)
but still not necessary. Be creative too!
Do not put [motion caption] in the same round, the user can also give motion to AI to reason from it too. 
Also, do not directly put [motion caption] twice in the round. You should put in only once, regarding both User and AI.
[motion caption] are motion strings with skeleton information, which are used to generate motion. Do not repeat the caption.
If you want to refer to these motions, just refer to it as the 'first motion'. But this motion string should be contained in the former to refer to.
Try to make [motion caption] in details that do not require the previous context to generate the motion physically.
** Instead of the user fully describing what to do next, be more implicit, especially for the second motion, focusing more on the story. **
questions-answers not limited to the above examples.
Questions should not be yes-no questions but wh-questions.
The User-AI round should design at most 2. [motion caption] should appear only twice.
Do not generate any new objects. Please follow the template from the example.
}
\greybox{
It is better to keep the questions and answers concise.
Try to be rational and keep in mind to make everything in sense, and the story smooth enough.

Do not mention facial expressions or hands. Make the [motion caption] only "twice" in the conversation.
[motion caption] should always contain a description of two people. 
[motion caption] should have enough details for the motion, letting the model generate a correct motion by only accessing this caption without the previous context. 
Do not change the style of the motion caption.
Do not make big and sudden changes in scenarios.
REMEMBER: Try to make a description of the second motion that can be inferred by seeing the first motion. 
DO NOT GENERATE conversations that can be understandable without the previous context. 
FOCUS on **editing** the motion based on the emotion or personas. 
Users should NEVER ask AI to generate the motion giving details about what to do. LET AI infer about what to do based on the change of emotion.
t is better to keep the questions and answers concise, with strictly following the format. Do not explain too much when generation motion.
You are making a conversation about how the motion of the one person will change based on the persona, instead of keeping the story going on.
The motion should be changed via body movement, not with facial expressions or hands.
Do not directly [motion caption], this is just the format to guide you to fill the description there. Strictly follow the format.
Generating **two** captions, with the changing persona for the motion. For the second caption, just change the motion of the second person.
Do NOT LEAVE the [motion caption] holder!
Do not put something like slightly, small, etc. It won't be able to be visualized!
Try to make a [motion caption] with the change of meaning of the motion, while maintaining a high-level scenario. Try to change the motion of the person dramatically, instead of changing just a few words.
}
Action labels contain all the action labels in the dataset, which bounds the captions to be inside the trained data from the text-to-motion model.

Next, prompts for motion reasoning and story generation without caption sample is as follows: 
\greybox{
You are an AI visual assistant, and you are seeing a motion.
Design a conversation between you and a person building a conversation about reasoning this motion.
In conversations, you should indicate who said using "User:", "AI:" in the beginning but these two words
do not occur in sentences.
The answers should be in a tone that an AI assistant is seeing the motion and answering the
question. The scenario should always contain two people in the scene.
Generate a conversation about building a story from two different motions. 
The flow of the conversation is as follows:

1. Creating a scenario. REMBER to make a story in this.
2. Reason about the motion or generate motion caption based on the scenario

"""Example:
    User: The current scene is [Two individuals sitting across from each other, with one person extending his/her left hand and the other person extending their left hand. They proceed to participate in a wrist-wrestling competition]. Can you tell me what happened before? 
    AI: Two people are doing arm-wrestling match, before that, two people will be doing fist dumps for fair play.
    User: Show me what will happen after that in motion format.
    AI: [One person is conducting a v-sign while the other stands still.]""",
"""Example:
    User: Two friends are doing an arm-wrestling match, show me the motion of that.
    AI: [Two individuals sit across from each other, with one person extending his/her left hand and the other person extending left hand. They proceed to participate in a wrist-wrestling competition]
    User: Show me what happened before that in motion format.
    AI: [two people are doing fist dumps].
    User: Why are they doing the fist dumps?
    AI: They are exchanging fist dumps to play a fair game in arm-wrestling.""",
"""Example:
    User: The current scene is [Two individuals sitting across from each other, with one person extending his/her left hand and the other person extending their left hand. They proceed to participate in a wrist-wrestling competition]. Can you tell me what happened before?. 
    AI: Two people are doing arm-wrestling match, before that, two people will be doing fist dumps for fair play.
    User: The next scene is [One person is conducting a v-sign while the other stands still.]. Explain the reason for this motion.
    AI: After the arm-wrestling match, one person won the game. The person is showing this happiness to the audience."""

===========
Example format for the [motion caption]:
    - One person approaches, raises his/her right hand to grab the other person's right forearm, places his/her left hand on it, and walks in the direction the grabbed person is facing.
    - Two people face each other, one person lifts his/her right leg and walks towards the other person, stopping half a meter away.
    - A person falls and braces himself/herself on the ground with his/her right hand. Another person approaches, squats down, and grabs his/her left arm with both hands to assist him/her in standing up.
The content inside the bracket ([]) is a caption for the motion. This is for visualizing the motion, which is not given in textual form during inference. 
I will denote this as [motion caption]. 
    
Please denote [motion caption] when AI or the user has to answer in the motion sequence.
Please make [motion caption] that is similar to the following action labels: \vio{[Action LABELS]}, and other motions like everyday routines (e.g., passing objects, greeting, communicating, etc.), and professional motions (e.g., Taekwondo, Latin dance, boxing, etc.)
}
\greybox{
but still not necessary. Be creative too!
Do not put [motion caption] in the same round, the user can also give motion to AI to reason from it too. 

Also, do not directly put [motion caption] twice in the round. You should put in only once, regarding both User and AI.
[motion caption] are motion strings with skeleton information, which are used to generate motion. Do not repeat the caption.

If you want to refer to these motions, just refer to it as the 'first motion'. But this motion string should be contained in the former to refer to.
Try to make [motion caption] in details that do not require the previous context to generate the motion physically.
** Instead of the user fully describing what to do next, be more implicit, especially for the second motion, focusing more on the story. **
questions-answers not limited to the above examples.
Questions should not be yes-no questions but wh-questions.
The User-AI round should design at most 2. [motion caption] should appear only twice.
Do not generate any new objects. Please follow the template from the example.
It is better to keep the questions and answers concise.
Try to be rational and keep in mind to make everything in sense, and the story smooth enough.
Do not mention facial expressions or hands. Make the [motion caption] only "twice" in the conversation.
[motion caption] should always contain a description of two people. 
[motion caption] should have enough details for the motion, letting the model generate a correct motion by only accessing this caption without the previous context. 
Do not make the conversation more than three rounds. 
}

Using the sample from the prior dataset, we have prompted the sampled motion and its corresponding caption to generate a multi-turn conversation that contains the sample motion. 
For motion reasoning and story generation tasks, we have prompted a large language model to generate a second motion caption and corresponding conversational data. Prompts are as follows: 
\greybox{
You are an AI visual assistant, and you are seeing a motion.
Design a conversation between you and a person building a conversation about reasoning this motion.
In conversations you should indicate who said using "User:", and" AI:" in the beginning but these two words
do not occur in sentences.
The answers should be in a tone that an AI visual assistant is seeing the motion and answering the
question. The scenario should always contain two people in the scene.
Generate a conversation about building a story from two different motions. 
The flow of the conversation is as follows:
1. Creating a scenario. REMBER to make a story in this.
2. Reason about the motion or generate motion caption based on the scenario
=====================
Motion 1:[Two individuals sit across from each other, with one person extending his/her left hand and the other person extending left hand. They proceed to participate in a wrist-wrestling competition]
"""Example:
    User: The current scene is [motion\_placeholder\_1]. Can you tell me what happened before? 
    AI: Two people are doing arm-wrestling match, before that, two people will be doing fist dumps for fair play.
    User: Show me what will happen after that in motion format.
    AI: [One person is conducting a v-sign while the other stands still.]""",
"""Example:
    User: Two friends are doing an arm-wrestling match, show me the motion of that.
    AI: [motion\_placeholder\_1]
    User: Show me what happened before that in motion format.
    AI: [two people are doing fist dumps].
    User: Why are they doing the fist dumps?
    AI: They are exchanging fist dumps to play a fair game in arm-wrestling.""",
    
"""Example:
    User: The current scene is [motion\_placeholder\_1]. Can you tell me what happened before?. 
    AI: Two people are doing arm-wrestling match, before that, two people will be doing fist dumps for fair play.
    User: The next scene is [One person is conducting a v-sign while the other stands still.]. Explain the reason for this motion.
    AI: After the arm-wrestling match, one person won the game. The person is showing this happiness to audience.""",
=====================
lease denote [motion\_placeholder] is when AI or the user has to answer in the motion sequence.
Example format for the [motion caption]:
    - One person approaches, raises his/her right hand to grab the other person's right forearm, places his/her left hand on it, and walks in the direction the grabbed person is facing.
    - Two people face each other, one person lifts his/her right leg and walks towards the other person, stopping half a meter away.
    - A person falls and braces himself/herself on the ground with his/her right hand. Another person approaches, squats down, and grabs his/her left arm with both hands to assist him/her in standing up.
The content inside the bracket ([]) is a caption for the motion. This is for visualizing the motion, which is not given in textual form during inference. 
I will denote this as [motion caption]. 
Please denote [motion caption] when AI or the user has to answer in the motion sequence.
Please make [motion caption] that is similar to the following action labels: \vio{[Action LABELS]}, and other motions like everyday routines (e.g., passing objects, greeting, communicating, etc.), and professional motions (e.g., Taekwondo, Latin dance, boxing, etc.)
but still not necessary. Be creative too!
!! Motion 1 is the description of [motion\_placeholder\_1]. Do not generate as [motion caption] for the first motion, rather just use [motion\_placeholder\_1].
DO NOT REPAT the given description, just use the [motion\_placeholder\_1]
For the second motion, make it as [description of motion that you want].
[motion caption] should always contain a description of two people. 
[motion caption] should have enough details for the motion, letting the model generate a correct motion by only accessing this caption without the previous context. 
Do not make the conversation more than three rounds.
Strictly follow the format of the given example. But not the motion inside there be creative. 
=====================
Motion1:\vio{[Motion caption from prior dataset]}
}

For the motion editing task, we have divided prompts into two parts. We first generate an edited motion caption with reasoning steps by prompting the large language model as follows:

\greybox{
First, let's edit the motion description. The provided motion descriptions represent the same motion. 
The motion content you are seeing is provided as follows:
Motion1:
\vio{Motion caption from prior dataset}
Focus on editing the motion based on the emotion, or based on persona like relationship or personality.
Remember that you cannot edit the motion related to face or hands. Just edit the body motion.
**Do not put something like slightly, small, etc. It won't be able to be visualized!**
Try to make a the meaning of the motion, while maintaing high-level scenario.
Format:
    Motion 2: []
Do not put adjective in new motion description, description would be about the movement without any styles of motion.
Instead of changing the style or size of the motion description, always change the motion itself that has different meaning.
Just generate it based on choosing one of the motion description, not all of them.
Try to change the motion of the person dramatically, instead of changing just few words.
But still maintain the high-level action label of this motion. DO not change the whole scenario.
}

Based on this generated edited motion caption and corresponding reasoning steps are then conditioned to the next prompts to generate the conversational data. 
\greybox{
You are an AI visual assistant, and you are seeing a motion.
Design a conversation between you and a person building a conversation about editing this motion.
In conversations, you should indicate who said using "User:", and "AI:" in the beginning but these two words
do not occur in sentences.
The answers should be in a tone that an AI visual assistant is seeing the motion and answering the
question. The scenario should always contain two people in the scene.
Generate a conversation about editing the motion based on two different given motions. 
The flow of the conversation is as follows:
1. Creating a scenario. 
2. Change the emotion or persona of just one person.
3. Describe how the motion will be changed.
=====================
Motion 1: [Two individuals sit across from each other, with one person extending his/her left hand and the other person extending both hands. They proceed to participate in a wrist-wrestling competition, where the second person utilizes both hands in an attempt to defeat the first person's left hand.].
Motion 2: [They sit across from each other, with one person extending his/her left hand and the other person extending both hands. They proceed to participate in a wrist-wrestling competition].
    """Example:
    User: Let's create a story starting from [motion\_placeholder\_1]. 
    AI: The one person kept losing the game, which made him competitive to win the game, like using his/her hands. 
    User: The next scene is [motion\_placeholder\_2].
    AI: Now, the person got a warning from the referee, leading him/her to just use one hand.""",
"""Example:
    User: Two friends are doing an arm-wrestling match.
    AI: [motion\_placeholder\_1]
    User: Okay one person looks too competitive in there. Can you make one person have more sportsmanship?
    AI: [motion\_placeholder\_2].
    User: Explain the reason for the motion.
    AI: One person may have gotten a warning from the referee..""",
"""Example:
    User: Two friends are doing an arm-wrestling match, like [motion\_placeholder\_1]. 
    AI: Two people are doing an arm-wrestling match, while one person is grabbing the other's left hand, one person is using both hands. 
    User: Okay one person looks too competitive in there. Can you make one person have more sportsmanship?
    AI: [motion\_placeholder\_2]""",
"""Example:
    User: Let's start making a story. Two friends are doing an arm-wrestling match, like [motion\_placeholder\_1]. 
    AI: The other person got a warning from the referee, leading him/her to just use one hand.
    User: Sounds interesting. Can you visualize it?
    AI: [motion\_placeholder\_2]"""
=====================
Please denote [motion\_placeholder] when AI or the user has to answer in the motion sequence.
[motion\_placeholder\_1] denotes Motion1, [motion\_placeholder\_2] denotes Motion2. Just use this term.
Do not put [motion\_placeholder]s in the same round, the user can also give motion to AI to reason from it too. 
Always follow the flow that motion 1 comes first.
If you want to refer to these motions, just refer to it as the 'first motion'. But this motion string should be contained in the former to refer to.
questions-answers not limited to the above examples.
** Instead of the user fully describing what to do next, be more implicit, especially for the second motion. **
questions-answers not limited to the above examples.
Questions should not be yes-no questions but wh-questions.
The User-AI round should design at most 2. 
Do not generate any new objects. Please follow the template from the example.
It is better to keep the questions and answers concise.
Try to be rational and keep in mind to make everything in sense.
Do not mention facial expressions or hands.
Do not make a big and sudden change in scenarios.
REMEMBER: Try to make a description of the second motion that can be inferred by seeing the first motion. 
DO NOT GENERATE conversations that can be understandable without the previous context. 
FOCUS on **editing** the motion based on the emotion or personas. 
Users should NEVER ask AI to generate the motion giving details about what to do. 
LET AI infer about what to do based on the change of emotion. 
**Focus on the change of persona.**
Strictly follow the format of the given example.
Put [motion\_placeholder\_1] and [motion\_placeholder\_2] each once in total conversation.
The motion content you are seeing is provided as follows:
Motion1:
\vio{Motion caption from prior dataset}
Motion2:
\vio{Generated Motion caption}
}
\clearpage
\section{Prompts for LLM-Assisted Evaluation}\label{sec:19}
To evaluate the reasoning ability of the proposed method, we have utilized LLM-assisted evaluation as shown in Section 5.2. The prompts used to evaluate such ability is as follows: 

\greybox{
We are evaluating the results of a model designed for generating interleaved motion-text documents. 
The model's input, starting with "INPUT:", can either be the beginning of a text-motion interleaved document or a specified topic. 
Its output, starting with "OUTPUT:", will then be either a continuation of the document or content generated based on the given topic. 
The motion is given as ground truth captions denoted as [c1, c2, c3] where all captions are describing the same motion.
Please remember that it is the caption of the motion, while there are many ways to describe the same motion. The provided caption is just part of it.
As an expert in multimodal evaluation, your task is to assess the quality of the output that is describe as text.

Scoring Guidelines:

- 0-3: Major deficiencies, misalignment, or inconsistency

- 4-7: Minor gaps, misalignment, or inconsistency

- 8-10: Complete and thorough alignment, strong consistency

Scoring Criteria:

1. Logical Coherence:

    - Evaluates the logical consistency and reasoning accuracy of the generated text
    
    - Key Aspects:
    
\hspace{0.5 cm}- Causal Relationships: Are the cause-and-effect relationships in the story or reasoning clear and sensible?
        
\hspace{0.5 cm}- Temporal Consistency: Does the timeline of events flow logically, without jumps or anachronisms?
        
\hspace{0.5 cm}- Character and Event Consistency: Do the actions of characters or descriptions of events remain consistent throughout the text?
        
\hspace{0.5 cm}- Plausibility: Does the explanation or story feel plausible, given the context of the motion data?

2. Content Alignment

    - Evaluate how accurately the generated text reflects the context of the given motion data
    
    - Key Aspects:
    
\hspace{0.5 cm}- Relevance: Does the generated text accurately respond to the motion data, staying relevant to the scenario presented by the input?
        
\hspace{0.5 cm} - Accuracy: Are the details and context derived from the motion data correctly reflected in the text?
        
\hspace{0.5 cm} - Interpretation: Does the text offer a reasonable interpretation or explanation of the motion, fitting within the implied scenario?

3. Naturalness:
   - Evaluate the quality of the output texts
   
    - Key Aspects:

\hspace{0.5 cm}- Fluency: Is the text grammatically correct, with smooth sentence structures?
        
\hspace{0.5 cm}- Readability: Does the text flow well, without awkward phrasing or confusing syntax?
        
\hspace{0.5 cm}- Tone and Style: Is the tone appropriate for the context? Does it match human-like writing in terms of style and nuance?
        
\hspace{0.5 cm}- Engagement: Is the text engaging and interesting to read?

JSON Output Structure: 

\{ 

\hspace{0.5 cm}    "scores": \{

\hspace{1 cm}        "Logical Coherence": \{

\hspace{1.5 cm}            "Justification": "brief justification of any deficiencies in image quality",

\hspace{1.5 cm}            "Score": 0-10
        \},

\hspace{1 cm}        "Content Alignment":\{

\hspace{1.5 cm}             "Justification": "brief justification of any deficiencies in image quality",

\hspace{1.5 cm}             "Score": 0-10 \},

\hspace{1 cm}         "Naturalness":\{

\hspace{1.5 cm}             "Justification": "brief justification of any deficiencies in image quality",

\hspace{1.5 cm} "Score": 0-10
        \}
     
\hspace{0.5 cm}    \}

\}

Data to Review: 
}
\end{document}